\documentclass[11pt]{article}

\usepackage[preprint]{acl}

\usepackage[T1]{fontenc}
\usepackage[utf8]{inputenc}
\usepackage{times}
\usepackage{latexsym}
\usepackage{microtype}
\usepackage{inconsolata}
\usepackage{amsmath,amsfonts,amssymb}
\usepackage{algorithm}
\usepackage[noend]{algpseudocode}
\usepackage{array}
\usepackage{booktabs}
\usepackage{bbding}
\usepackage{enumitem}
\usepackage{graphicx}
\usepackage{longtable}
\usepackage{makecell}
\usepackage{multirow}
\usepackage{nicefrac}
\usepackage{tabularx}
\usepackage[most]{tcolorbox}
\usepackage{textcomp}
\usepackage{url}
\usepackage{verbatim}
\usepackage{wasysym}
\usepackage{wrapfig}
\usepackage{pifont}
\usepackage[table,dvipsnames]{xcolor}

\newtcolorbox{insightbox}{
  colback=gray!10,
  colframe=gray!60,
  boxrule=0.5pt,
  arc=2pt,
  left=6pt,
  right=6pt,
  top=4pt,
  bottom=4pt
}

\newcommand{\cmark}{\ding{51}}
\newcommand{\xmark}{\ding{55}}
\newcommand{\pmark}{$\blacktriangle$}
\newcolumntype{C}[1]{>{\centering\arraybackslash}p{#1}}

\title{ Can LLM Agents Sustain Long-Horizon Organizational Dynamics?}

\author{
\begin{tabular}{c}
Xuancheng Zhu\textsuperscript{1},
Yang Yue\textsuperscript{1},
Shuaibing Wan\textsuperscript{1},
Zihan Dou\textsuperscript{1}, \\
Xiaohan Zhang\textsuperscript{1},
Yongrui Liu\textsuperscript{1},
Guoshun Nan\textsuperscript{1} \\
\small \textsuperscript{1}Beijing University of Posts and Telecommunications, Beijing, China
\end{tabular}
}

\begin{document}
\maketitle

\begin{abstract}
Large language agents are increasingly used for social simulation, yet it remains unclear whether they can sustain coherent behavior in structured organizations, where goals must propagate through hierarchy, tasks depend on prior execution, and artifacts accumulate over long horizons. We formulate long-horizon organizational simulation as a memory-centered coordination problem and introduce TaskWeave, a hierarchical agentic framework that maintains planning states through a Formulate–Partition–Diagnose–Align cycle and grounds execution through dependency-aware trace memory.   We evaluate TaskWeave in a year-long IT company simulation and compare it with other multi-agent frameworks on organizational coherence, execution grounding, and downstream enterprise NLP utility. Experiments show that TaskWeave supports coherent and long-horizon organizational dynamics while producing grounded artifacts and adapting to external environments. These findings suggest that structured simulation memory is a key mechanism for building reliable LLM-based organizational simulators.
\end{abstract}

\section{Introduction}

Large language model (LLM) agents are increasingly used as simulators of social, economic, and interactive systems~\cite{park2023stanfordtown,li-etal-2024-econagent,yang2025twinmarket,xu2023urbangenerative,gao2023s3,yukhymenko2024a}.
By endowing language models with roles, memory, tools, and interaction protocols, recent agentic systems can generate adaptive behaviors beyond single-turn task solving~\cite{cao2024survey,chen2024survey,yehudai2025surveyevaluationllmbasedagents,openclaw2026github}.
This progress makes it possible to study complex human-centered systems in controllable environments, without relying on costly, private, or difficult-to-repeat real-world experiments~\cite{vezhnevets2023generative}.

Despite this progress, structured organizations remain a challenging and underexplored setting for LLM-agent simulation.
Organizations are not merely collections of interacting individuals, nor are they static workflows~\cite{teece2007explicating,zollo2016toward}.
They involve hierarchical authority, role-specific responsibilities, long-horizon goals, interdependent tasks, evolving external conditions, and accumulated process artifacts.
These properties distinguish organizational simulation from open-ended social interaction and conventional business process modeling~\cite{van2011process,dumas2018fundamentals,rolon2012agent}.
A useful organizational simulator must therefore preserve not only locally plausible actions, but also the evolving organizational state behind them.

\begin{figure*}[t]
  \centering
  \includegraphics[width=\textwidth,trim=5 0 5 0,clip]{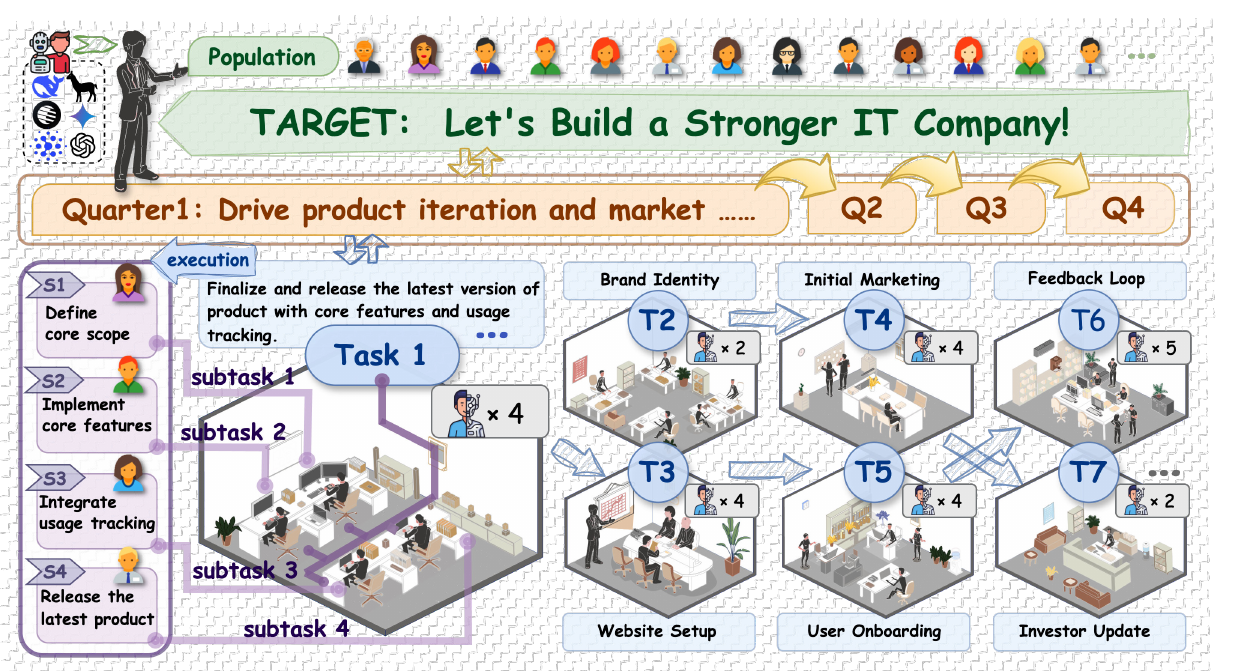}
  \caption{Overview of organizational dynamics. High-level goals are progressively refined into interdependent tasks, coordinated across hierarchical roles, and executed by agents under evolving contexts. The key challenge is to maintain coherent planning and execution states over long horizons.}
  \label{fig:intro}
  \vspace{-15pt}
\end{figure*}

We refer to this challenge as the \emph{long-horizon organizational coherence problem}.
Existing LLM-agent frameworks provide useful mechanisms for multi-agent collaboration~\cite{yun2026graphofagents}, tool use~\cite{wu2023autogen}, and autonomous execution~\cite{hou2024coactgloballocalhierarchyautonomous,cho2026safetyagent}, but they are not primarily designed to maintain organization-level state across hierarchy, time, and execution dependencies.
As a result, agents may generate locally reasonable outputs while gradually losing consistency in role assignment, task dependencies, plan inheritance, or execution history.
For organizational simulation, such failures are critical because later decisions must remain consistent with earlier plans, artifacts, and evidence.

More concretely, organizational simulation requires coherence along three coupled dimensions.
First, \textit{collaboration dynamics} require agents to coordinate through explicit roles, delegation paths, and inter-agent dependencies rather than unconstrained dialogue~\cite{galbraith2014designing}.
Second, \textit{planning dynamics} require high-level objectives to be recursively refined into temporally layered plans while preserving intent across levels~\cite{xu2008approach,hou2024coactgloballocalhierarchyautonomous}.
Third, \textit{execution dynamics} require actions to be grounded in prior outputs, intermediate progress, and external evidence rather than generated as isolated responses~\cite{kumar2026evolving}.
Together, these requirements turn long-horizon organizational simulation into a \emph{memory-centered coordination problem}: agents must act over a shared organizational state that evolves across plans, tasks, and execution traces.

To address this problem, we propose \textit{TaskWeave}, a hierarchical agentic framework for sustaining long-horizon organizational dynamics.
TaskWeave models an organization as a role-structured agent population constrained by an explicit delegation topology.
On top of this organizational prior, TaskWeave introduces a \emph{hierarchical simulation-memory mechanism} that maintains planning and execution states over time.
The planning side uses a multi-level \emph{Formulate--Partition--Diagnose--Align} (FPDA) cycle to propagate intent from high-level goals to localized task bundles, incorporate execution feedback, and update planning states.
The execution side uses dependency-aware operational execution to resolve prior results and external evidence before assigning and completing tasks.
Outputs are written back as traceable process memory, enabling later planning and execution to reuse earlier organizational context.
We evaluate TaskWeave in a year-long IT-company simulation and compare it with representative multi-agent frameworks.

\textbf{Our contributions are summarized as follows.}
\textbf{(i)} We formulate long-horizon organizational simulation as a memory-centered coordination problem, emphasizing role consistency, intent propagation, dependency satisfaction, and traceable execution.
\textbf{(ii)} We propose \textit{TaskWeave}, a hierarchical agentic framework that combines organizational priors, FPDA-based planning-state memory, and dependency-aware trace memory.
\textbf{(iii)} We conduct empirical evaluation in a year-long organizational simulation, showing that TaskWeave improves organizational coherence, grounds execution in prior context, and produces process-grounded artifacts useful for downstream enterprise NLP analysis.
\section{Related Work}
\label{sec:related_work}

Organizational processes have long been modeled through \emph{Business Process Management} (BPM), including BPMN-based workflows~\cite{van2011process,dumas2018fundamentals}, process redesign~\cite{reijers2003redesign}, declarative modeling~\cite{PesicDECLARE}, process mining~\cite{aalst2013business}, and rule-based agent simulation~\cite{rolon2012agent}. These methods support well-defined procedures, but rely on static templates and rules~\cite{augusto2018automated}, limiting their ability to model evolving enterprise processes.

LLM-based agentic systems extend simulation with reasoning, planning~\cite{hu2024agentgenenhancingplanningabilities}, memory~\cite{gao2024agentscope}, tool use~\cite{hong2024metagpt}, and communication~\cite{park2023stanfordtown}. They have been applied to collaborative task solving~\cite{shenSmallLLMsAre2024,duImprovingFactualityReasoning2023,li2025robust}, social and economic simulation~\cite{gao2023s3,yang2025twinmarket,li-etal-2024-econagent}, domain-specific settings~\cite{liAgentHospitalSimulacrum2024,wangUserBehaviorSimulation2024,xu2025theagentcompany}, and computer-use environments~\cite{xie2024osworld,openclaw2026github,zhaoCompeteAIUnderstandingCompetition2024,gao2025multimodal}. However, most target specific tasks~\cite{cao2025exploring}, short-horizon collaboration~\cite{ding2023multiagent}, or open-ended interaction, rather than hierarchical organizational processes with persistent planning--execution dependencies. TaskWeave addresses this gap through memory-centered coordination across temporal.

\section{Method}

\begin{figure*}[t]
  \centering
  \includegraphics[width=\textwidth]{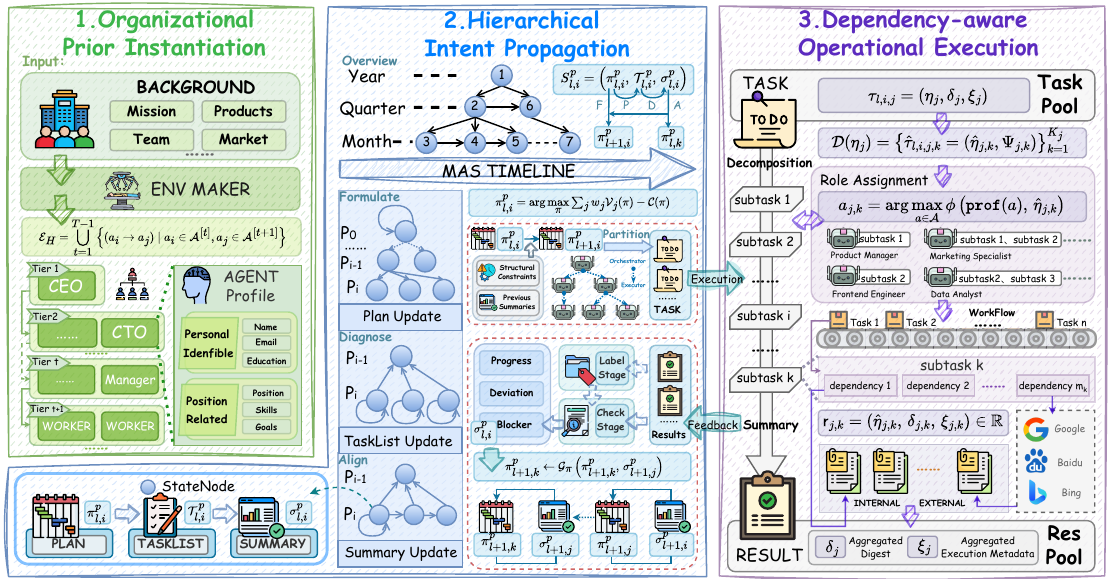}
  \caption{Overview of TaskWeave. Organizational metadata is compiled into a role-structured prior. A hierarchical simulation-memory mechanism then sustains organizational dynamics through FPDA-based planning-state propagation and dependency-aware operational execution. Execution outputs are written back as traceable process memory, forming a closed-loop simulation process.}
  \label{fig:main}
  \vspace{-10pt}
\end{figure*}

TaskWeave is a hierarchical agentic framework for simulating long-horizon organizational dynamics.
Its central design is a \emph{hierarchical simulation-memory mechanism}: planning states preserve organizational intent and task commitments across temporal levels, while execution states preserve grounded outputs and provenance for later reuse.
As shown in Figure~\ref{fig:main}, TaskWeave consists of three components.
First, \textbf{Organizational Prior Instantiation} compiles organizational metadata into a role-structured agent population and delegation topology.
Second, \textbf{Hierarchical Intent Propagation} maintains planning states through a multi-level Formulate--Partition--Diagnose--Align (FPDA) cycle.
Third, \textbf{Dependency-Aware Operational Execution} grounds task execution in role constraints, resolved dependencies, external evidence, and traceable result memory.

\subsection{Organizational Prior Instantiation}
\label{sec3.2}

TaskWeave begins by instantiating an \emph{organizational prior} from metadata, rather than treating agents as interchangeable workers.
Let $\mathbb{B}$ denote organizational metadata, including the domain, scale, departments, and role taxonomy of the target organization.
From $\mathbb{B}$, TaskWeave constructs an agent population $\mathcal{A}=\{a_i\}_{i=1}^{N}$, where each agent is represented as $a_i=(\phi_i,\rho_i)$.
Here, $\phi_i\sim\mathcal{P}_{\phi}(\mathbb{B})$ specifies personal attributes such as background and experience, while $\rho_i\sim\mathcal{P}_{\rho}(\mathbb{B})$ specifies the functional role description, including title, department, responsibility scope, and interaction privileges.
Thus, $\phi_i$ characterizes who the agent is, and $\rho_i$ constrains what the agent is expected to do.

To encode organizational structure, agents are partitioned into $T$ tiers, $\mathcal{A}=\bigcup_{t=1}^{T}\mathcal{A}^{[t]}$, where each tier $\mathcal{A}^{[t]}$ corresponds to a role family $\mathcal{R}^{[t]}$.
Fine-grained coordination is further captured by a delegation graph $\mathbb{H}=(\mathcal{A},\mathcal{E}_H)$, where each edge $(a_i,a_j)\in\mathcal{E}_H$ denotes a valid delegation or coordination path.
The edge set combines adjacent-tier delegation links and structurally permitted cross-unit links, i.e., $\mathcal{E}_H=\mathcal{E}_{\mathrm{tier}}\cup\mathcal{E}_{\mathrm{cross}}$.
This prior constrains how intent propagates during planning and which agents may receive tasks.
Therefore, $\mathbb{H}$ is not merely an organization chart, but the structural scaffold over which simulation memory flows.

\subsection{Hierarchical Intent Propagation}
\label{sec3.3}

Organizational dynamics emerge when high-level goals are recursively refined into local plans and executable tasks.
TaskWeave treats this process as hierarchical state maintenance.
A planning state is defined as
\begin{equation}
S_{l,i}^{p}=
\left(\pi_{l,i}^{p},\mathcal{T}_{l,i}^{p},\sigma_{l,i}^{p}\right),
\label{eq:planning_state}
\end{equation}
where $\pi_{l,i}^{p}$ is the current plan, $\mathcal{T}_{l,i}^{p}$ is the associated task bundle, and $\sigma_{l,i}^{p}$ is a diagnostic summary derived from downstream execution.
The index $l$ denotes a temporal level, such as year, quarter, month, or week; $i$ indexes a local planning unit within that level; and $p$ is a parent pointer identifying the upstream state from which the current state inherits intent.
The root state $S_{0,1}^{\varnothing}$ encodes the top-level organizational objective and initial environment.

This state representation separates three forms of simulation memory.
The plan $\pi$ records what the organization intends to achieve.
The task bundle $\mathcal{T}$ records what has been delegated or committed.
The diagnostic summary $\sigma$ records what has been observed from execution.
This separation allows TaskWeave to maintain intent, commitment, and feedback across long horizons.

To update planning states, TaskWeave uses the \textbf{Formulate--Partition--Diagnose--Align} (FPDA) cycle.
Given a parent state $S_{l,i}^{p}$ and a target tier $\mathcal{A}^{[t]}$, the transition is defined as
\begin{equation}
S_{l+1,j}^{i}
=
\mathcal{F}_{\mathrm{FPDA}}
\left(S_{l,i}^{p},\mathcal{A}^{[t]},\mathbb{H}\right).
\label{eq:fpda_transition}
\end{equation}
Here, $j$ indexes the child planning unit, and the superscript $i$ indicates that the child state inherits from parent unit $i$.
This transition updates the three components of the planning state.
\textbf{Formulate} derives a local plan from upstream intent and diagnostic context.
\textbf{Partition} converts the local plan into a task bundle that can be delegated through $\mathbb{H}$.
\textbf{Diagnose} summarizes execution results, deviations, and blockers from the result pool.
\textbf{Align} revises the local plan using diagnostic feedback.

Concretely, starting from $S_{l,i}^{p}=(\pi,\mathcal{T},\sigma)$, FPDA first formulates a local plan
$\pi'=\mathcal{F}_{\mathrm{form}}(\pi,\sigma,\mathbb{B})$,
then partitions it into executable tasks
$\mathcal{T}'=\mathcal{F}_{\mathrm{part}}(\pi',\mathcal{A}^{[t]},\mathbb{H})$.
The produced bundle $\mathcal{T}'$ becomes the executable task set associated with the child planning state.
After execution, TaskWeave derives a diagnostic summary
$\sigma'=\mathcal{F}_{\mathrm{diag}}(\mathbb{R},\mathcal{T}')$
and obtains the aligned plan
$\tilde{\pi}'=\mathcal{F}_{\mathrm{align}}(\pi',\sigma')$.
The resulting child state is
\begin{equation}
S_{l+1,j}^{i}
=
(\tilde{\pi}',\mathcal{T}',\sigma').
\label{eq:fpda_child_state}
\end{equation}
Here, primes denote updated local state components.
In this way, FPDA propagates intent downward, incorporates execution evidence upward, and keeps planning states coherent across temporal and organizational layers.

TaskWeave also models cross-unit synchronization and upward feedback.
When a state produces diagnostic feedback, sibling states in $\mathcal{N}_{\mathrm{sibling}}(j)$, i.e., states sharing the same parent as state $j$, may update their plans through an alignment operator $\mathcal{G}_{\mathrm{align}}$, capturing coordination such as deadline adjustment, budget negotiation, or interdepartmental dependency resolution.
Execution summaries are then aggregated upward through $\mathcal{U}_{\mathrm{up}}(\{\sigma_{l+1,j}^{i}\}_{j\in\mathcal{C}(i)})$, where $\mathcal{C}(i)$ denotes the children of planning unit $i$.
Here, $\mathcal{G}_{\mathrm{align}}$ and $\mathcal{U}_{\mathrm{up}}$ are abstract update operators implemented by structured prompts.
% Together, downward intent propagation, lateral synchronization, and upward feedback form the planning-side memory update.

\subsection{Dependency-Aware Execution}
\label{sec3.4}

TaskWeave realizes propagated task bundles as grounded actions.
Rather than treating execution as stateless generation, TaskWeave conditions each action on role constraints, resolved dependencies, accumulated process memory, and external evidence.
For a composite task in state $S_{l,i}^{p}$, we define
\begin{equation}
\tau_{l,i,m}=(\eta_m,\delta_m,\xi_m)\in\mathcal{T}_{l,i}^{p},
\label{eq:composite_task}
\end{equation}
where $m$ indexes a composite task, $\eta_m$ is the task specification, $\delta_m$ is the output field, and $\xi_m$ stores provenance metadata.
The task specification $\eta_m$ contains the background, task description, constraints, and priority, while $\delta_m$ and $\xi_m$ are filled after execution.

TaskWeave decomposes each composite task into atomic subtasks $\{\hat{\tau}_{m,k}\}_{k=1}^{K_m}$, where each subtask is represented as $\hat{\tau}_{m,k}=(\hat{\eta}_{m,k},\Psi_{m,k})$.
Here, $k$ indexes an atomic subtask, $\hat{\eta}_{m,k}$ is the atomic goal, and $\Psi_{m,k}$ is a dependency query specifying what information must be resolved before execution.
A dependency query may request prior internal results, external evidence, or both.
Its resolved context is
\begin{equation}
\Gamma_{m,k}=
\mathcal{Q}(\Psi_{m,k},\mathbb{R},\Omega)
=
\Gamma_{m,k}^{\mathrm{int}}
\cup
\Gamma_{m,k}^{\mathrm{ext}},
\label{eq:dependency_resolution}
\end{equation}
where $\mathbb{R}$ is the global result pool, $\Omega$ is the accessible tool space, $\Gamma_{m,k}^{\mathrm{int}}$ is retrieved from prior organizational results, and $\Gamma_{m,k}^{\mathrm{ext}}$ is obtained from tools or external environments.
Thus, $\Psi_{m,k}$ acts as a structured memory-addressing query over organizational memory and external evidence.

Each atomic subtask is then dispatched to a role-compatible agent.
Let $t$ be the target execution tier implied by the delegation path.
TaskWeave selects
\begin{equation}
a_{m,k}^{*}
=
\arg\max_{a\in\mathcal{A}^{[t]}:\,(a_{\mathrm{src}},a)\in\mathcal{E}_H}
s_{\mathrm{match}}(a,\hat{\eta}_{m,k}),
\label{eq:role_dispatch}
\end{equation}
where $a_{\mathrm{src}}$ is the delegating agent and $s_{\mathrm{match}}$ measures compatibility between the candidate agent's role description and the subtask semantics.
This ties execution to the organizational prior, ensuring that tasks are completed by agents with appropriate authority and expertise.

Given subtask $\hat{\tau}_{m,k}$, resolved context $\Gamma_{m,k}$, and assigned agent $a_{m,k}^{*}$, execution produces a subtask result
\begin{equation}
r_{m,k}
=
\mathcal{E}_{\mathrm{exec}}
(\hat{\eta}_{m,k},\Gamma_{m,k},a_{m,k}^{*}).
\label{eq:execution}
\end{equation}
Each result is represented as $r_{m,k}=(\hat{\eta}_{m,k},\delta_{m,k},\xi_{m,k})$, where $\delta_{m,k}$ is the generated content and $\xi_{m,k}$ records subtask-level provenance, including the responsible agent, referenced dependencies, tool interactions, and execution context.
After all atomic subtasks of $\tau_{l,i,m}$ are completed, their local results $\mathcal{B}_{m}=\{r_{m,k}\}_{k=1}^{K_m}$ are merged into a composite-level output $(\delta_m,\xi_m)=\mathcal{M}(\mathcal{B}_{m})$, which is then written back to the global result pool:
\begin{equation}
\mathbb{R}\leftarrow \mathbb{R}\cup\{(\eta_m,\delta_m,\xi_m)\}.
\label{eq:result_merge}
\end{equation}
Here, subscript $(m,k)$ denotes subtask-level outputs and provenance, while subscript $m$ denotes the merged composite-level result.

The result pool therefore functions as traceable process memory rather than passive storage.
It preserves composite outputs together with their dependency structure, responsible agents, tool evidence, and execution context.
Later tasks can retrieve this memory through dependency queries, and higher-level FPDA states can use it for diagnosis and alignment.
This closes the simulation loop: planning states generate tasks, execution grounds tasks in memory and evidence, and results are written back to support subsequent planning and execution.
\section{Experiments}
\label{sec:exp}

\subsection{Experimental Setup}
\label{sec:simulation}

We evaluate \textit{TaskWeave} through a year-long simulation of a software-as-a-service (SaaS) company, \textbf{Company A}, with \textbf{15 agents}, \textbf{3 hierarchical tiers} (\emph{boss}, \emph{managers}, and \emph{workers}), and \textbf{3 departments} (\emph{Technology}, \emph{Marketing}, and \emph{Strategy}).
The simulation follows a $\text{year} \rightarrow \text{quarter} \rightarrow \text{month} \rightarrow \text{week}$ planning hierarchy.
Unless otherwise specified, organizational metadata, role definitions, and task background are fixed across experiments.
We instantiate TaskWeave with six LLM backbones: \texttt{GPT-4o-mini}, \texttt{Gemini-2.0-Flash}, \texttt{Deepseek-v3}, \texttt{Moonshot-v1-8K}, \texttt{LLaMA3.1-70B}, and \texttt{GLM-4-Flash}.

The evaluation follows the TaskWeave simulation loop: organizational coherence tests role-consistent delegation, plan propagation tests intent preservation across temporal layers, grounded execution inspects reusable process traces, enterprise NLP utility probes generated artifacts, and responsiveness/transferability evaluates external coupling and cross-domain reuse.
We combine automatic metrics, structured LLM-based evaluation, and human auditing.
Detailed prompts, role definitions, reliability checks, and evaluation instructions are provided in Appendix~\ref{app:evaluation}.

\subsection{Organizational Coherence}
\label{sec:coherence}

A core requirement of organizational simulation is that work should be distributed according to role structure rather than collapsing onto a few frequently activated agents.
We therefore evaluate whether the organizational prior and delegation topology are reflected in execution.
We define an expert-designed reference role prior over the 11 executable worker roles in Company A, reflecting expected responsibility allocation in the simulated SaaS setting.
This prior is used to assess structural consistency rather than universal correctness.
We measure KL divergence against this prior and visualize workload proportions to expose model-specific delegation biases. (Detailed in App~\ref{app:role-distribution}.)

\begin{figure}[h]
  \centering
  \includegraphics[width=\columnwidth]{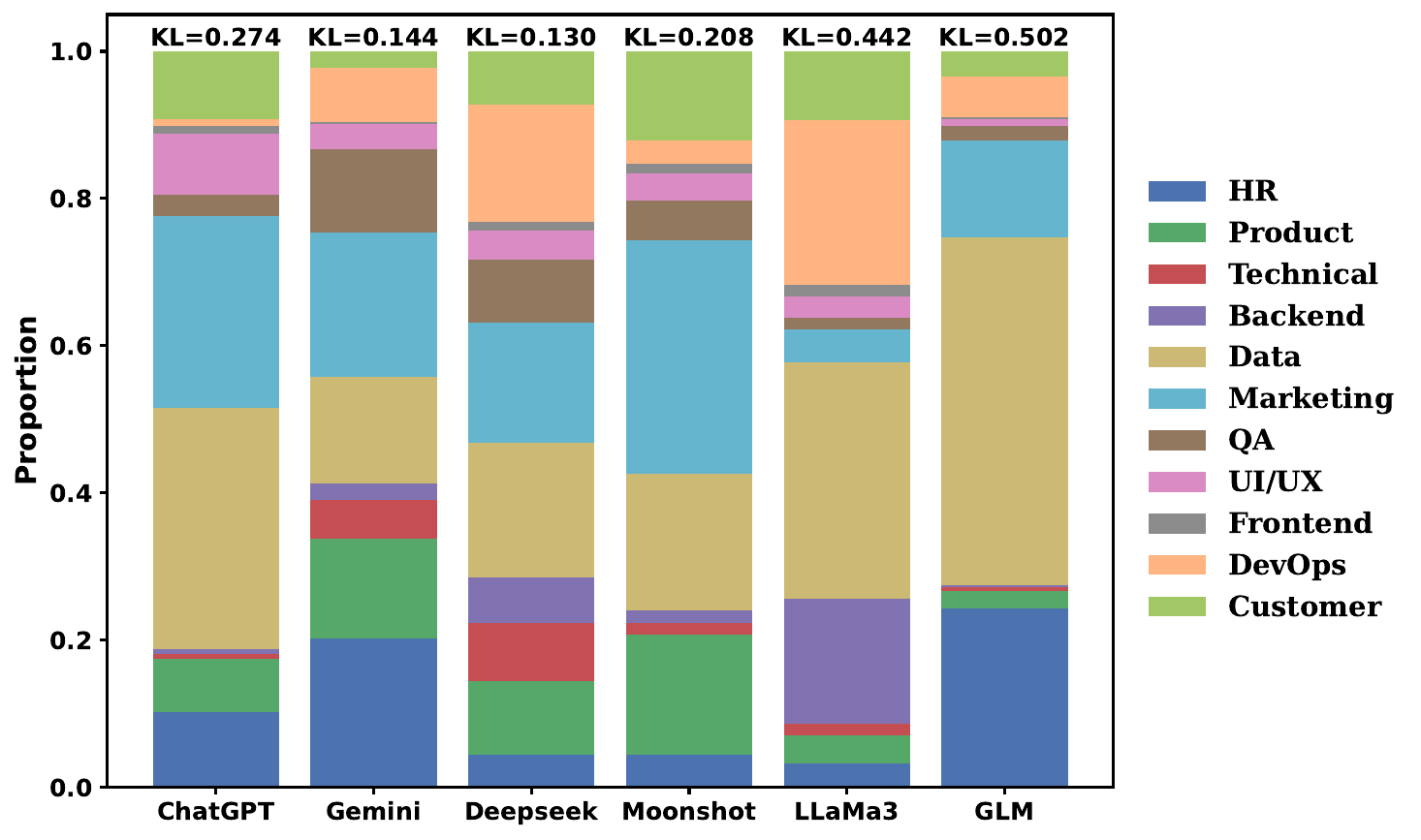}
  \caption{Role assignment distribution across different backbones. Each bar shows the proportion of tasks executed by each role, together with the KL divergence to the reference role prior.}
  \label{fig:role-dist}
  \vspace{-10pt}
\end{figure}

\noindent\textbf{Obs. \ding{172}: Organizational structure guides delegation, but model capacity determines whether roles are actually activated.}
Figure~\ref{fig:role-dist} shows that stronger backbones such as Gemini and Deepseek achieve broader role coverage and lower divergence, whereas weaker backbones such as GLM exhibit more skewed activation patterns.
Under less capable backbones, some roles are repeatedly overused while others remain under-activated, even though the orchestration mechanism permits broader delegation.
This suggests that role-grounded simulation depends not only on organizational constraints, but also on the backbone's ability to internalize and enact workload allocation.

\subsection{Plan Propagation}
\label{sec:PropagationDynamics}
We next examine whether high-level organizational plans can be propagated into lower-level executable work under FPDA.
For each monthly or weekly plan, an external evaluator model (\texttt{Gemini-2.5-Flash}) first generates a checklist of actionable checkpoints.
\textbf{n-check} denotes the average number of checkpoints used for evaluation.
We estimate completion rates from execution summaries under two criteria:
\textbf{Timely} completion requires a checkpoint to be fulfilled within the designated phase, while \textbf{Finalized} completion also credits delayed resolution in later cycles.
The gap between them captures temporal spillover, a common property of long-horizon organizational processes. Detailed evaluation protocols in App~\ref{app:MASCheckpoint}.

\begin{table}[h]
\caption{Completion rate across different backbones under monthly and weekly evaluation. ``-'' indicates that the model fails to produce sufficiently structured plans for stable checklist-based evaluation.}
\centering
\renewcommand{\arraystretch}{1.15}
\setlength{\tabcolsep}{3pt}
\resizebox{\columnwidth}{!}{%
\begin{tabular}{c|ccc|ccc}
\Xhline{1.2pt}
\rowcolor{CadetBlue!20}
\textbf{Model} & \multicolumn{3}{c|}{\textbf{Monthly}} & \multicolumn{3}{c}{\textbf{Weekly}} \\
\rowcolor{gray!10}
& \textbf{n-check} & \textbf{Timely} & \textbf{Finalized} & \textbf{n-check} & \textbf{Timely} & \textbf{Finalized} \\
\Xhline{1.2pt}
ChatGPT & 30 & 76.67\% & 91.39\% & 14.25 & 71.35\% & 95.32\% \\
\rowcolor{gray!20}
Gemini & 29.58 & 81.92\% & 92.66\% & 14.41 & 87.28\% & 98.58\% \\
Deepseek & 27.8 & 75.18\% & 82.01\% & 13.41 & 78.88\% & 92.55\% \\
\rowcolor{gray!20}
Moonshot & 24.3 & 71.03\% & 83.79\% & 13.08 & 72.61\% & 91.72\% \\
LLaMA3 & - & - & - & - & - & - \\
\rowcolor{gray!20}
GLM & - & - & - & 11.53 & 57.80\% & 68.21\% \\
\Xhline{1.2pt}
\end{tabular}%
}
\label{tab:llm_task_completion}
\end{table}

% \noindent\textbf{Obs. \ding{173}: Long-horizon simulation is limited less by generating plans than by preserving intent across time.}
% Table~\ref{tab:llm_task_completion} shows that stronger backbones progressively decompose and realize plans across monthly and weekly layers, while weaker backbones such as GLM and LLaMA3 fail to maintain stable propagation.
% Weekly goals are generally completed more reliably than monthly ones, reflecting the lower coordination burden and shorter dependency horizon at finer granularity.
% The gap between \textbf{Timely} and \textbf{Finalized} completion further reveals temporal carry-over: many objectives miss their intended window but are later resolved.
% This pattern indicates that TaskWeave can capture organizational spillover, while effective long-horizon propagation still depends on backbone reasoning quality.
\noindent\textbf{Obs. \ding{173}: Long-horizon simulation is constrained by intent preservation rather than plan generation alone.}
Table~\ref{tab:llm_task_completion} reveals a gap between short-range execution and long-range organizational continuity.
Stronger backbones maintain high finalized completion across monthly and weekly layers, whereas weaker backbones either fail to produce structured plans or lose stable propagation.
This suggests that the key difficulty is not generating plausible task lists, but preserving inherited intent as plans are decomposed, executed, and revised over time.
Weekly goals are easier because they involve shorter dependency chains, while monthly goals require longer memory over commitments, blockers, and delayed outcomes.
The gap between \textbf{Timely} and \textbf{Finalized} completion captures organizational spillover: objectives may miss their intended phase but remain recoverable in later cycles.
Thus, TaskWeave supports recoverable long-horizon plan continuity, while its effectiveness remains bounded by the backbone's ability to track intent and dependencies over time.

\begin{figure*}[t]
  \centering
  \includegraphics[width=\textwidth, trim=12 0 8 0, clip]{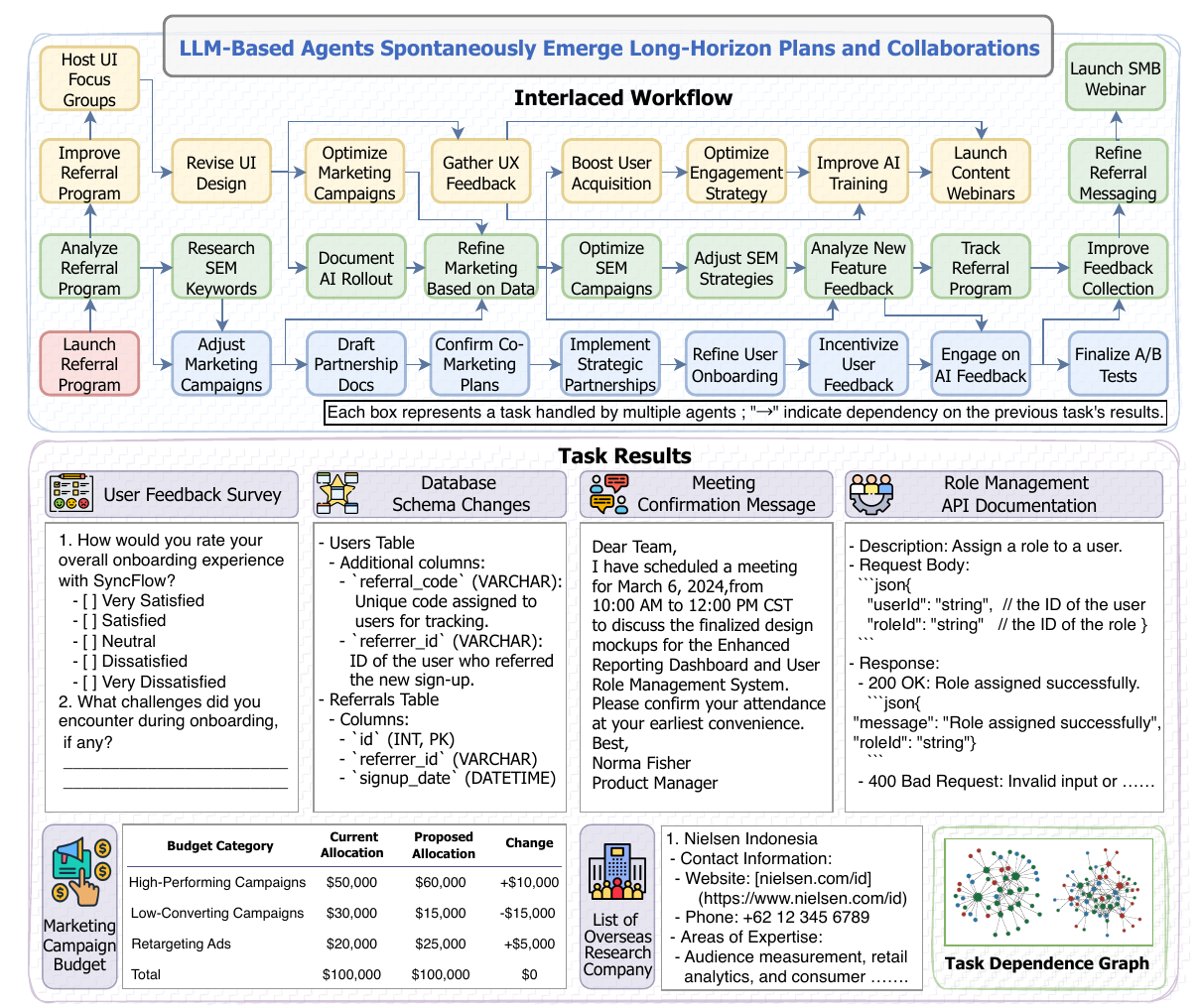}
  \caption{Case study. TaskWeave produces adaptive and temporally dependent workflows with diverse structured outputs; the dependency graph reveals inter-task relations and core coordination hubs.}
  \label{fig:case}
  \vspace{-15pt}
\end{figure*}

\subsection{Grounded Execution}
\label{sec:TaskExecution}

Beyond plan propagation, organizational simulation requires execution traces that remain anchored to prior process history, role constraints, and task dependencies.
We therefore inspect whether TaskWeave produces executable workflows whose intermediate artifacts can be reused by later tasks.

\noindent\textbf{Obs. \ding{174}: Grounded execution turns generated artifacts into reusable process memory.}
Figure~\ref{fig:case} presents a representative workflow generated without predefined scripts.
The workflow contains hierarchical task decomposition, temporal dependencies, and heterogeneous outputs, including plans, records, documents, and communication artifacts.
The dependency graph further exposes coordination hubs and inter-task links.
These traces show that TaskWeave does not produce isolated responses; instead, it links outputs through dependencies and provenance, making them reusable states for later planning and execution. The full ranked dependency graph and key high-degree tasks are reported in Appendix~\ref{app:KeyTasks}, showing that such reuse is not limited to the displayed case.

\subsection{Organizational NLP Utility}
\label{sec:Downstream}
Process-grounded organizational artifacts should contain structured, context-dependent enterprise information.
We instantiate this utility through \textbf{Organizational Sensitive Span Detection (OSSD)}, motivated by compliance needs such as GDPR~\cite{GDPR2016} and the EU Data Act~\cite{EUDataAct2023}.
Given a text $T$, OSSD detects sensitive spans with both category and rationale:
\[
S = \{(t_i,c_i,r_i)\}_{i=1}^{k},
\]
where $t_i$ is a span, $c_i$ is a label (e.g., \texttt{Financial Data}), and $r_i$ is a rationale.
Unlike standard named entity recognition, OSSD targets organization-specific, context-dependent spans and explicitly requires justifications.

\textbf{Annotation protocol.}
Following design principles for LLM-as-a-judge~\cite{yehudai2025surveyevaluationllmbasedagents}, we adopt a 3 stage \emph{label-and-verify} pipeline with \texttt{GPT-4o-mini}: span extraction, rationale generation, and verification.
The verification stage checks category consistency, rationale sufficiency, and span-boundary validity, and we manually audit a subset of outputs. (Detailed in App~\ref{app:OSSD}.)

\begin{table}[h]
\small
\centering
\renewcommand{\arraystretch}{1.2}
\setlength{\tabcolsep}{3pt}
\caption{Internal sensitive span quantity and usage under the shared single-stage organizational execution setting.}
\begin{tabular}{c|ccccc}
\Xhline{1.2pt}
\rowcolor{CadetBlue!20}
\textbf{Method} & \textbf{API} & \textbf{Spans} & \textbf{Len} & \textbf{Tokens (P/C)} & \textbf{Div.} \\
\Xhline{1.2pt}
AutoGen & 18.4 & 401 & 31.65 & 51k/8k & 14 \\
\rowcolor{gray!20}
MetaGPT & 14.7 & 528 & 36.91 & 56k/11k & 21 \\
CAMEL & 7.8 & 336 & 29.48 & 33k/6k & 12 \\
\rowcolor{gray!20}
MoA & 14.1 & 454 & 34.77 & 27k/5k & 11 \\
Magentic-One & 31.0 & 1538 & 52.04 & 465k/19k & 83 \\
\rowcolor{gray!20}
G-Designer & 40.6 & 252 & 28.22 & 16k/25k & 19 \\
Ours & \textbf{5.6} & \textbf{643} & \textbf{37.78} & \textbf{30k/4k} & \textbf{31} \\
\Xhline{1.2pt}
\end{tabular}
\label{tab:downstreamTask}
\vspace{-15pt}
\end{table}

\textbf{Evaluation setting.}
Since most baselines do not support the full year--quarter--month--week simulation loop, we compare them under a shared single-stage organizational execution setting.
This isolates whether structured execution produces richer enterprise-sensitive artifacts under comparable task inputs.
We compare against \texttt{AutoGen}~\cite{wu2023autogen}, \texttt{MetaGPT}~\cite{hong2024metagpt}, \texttt{CAMEL}~\cite{li2023camel}, \texttt{MoA}~\cite{wang2025mixtureofagents}, \texttt{Magentic-One}~\cite{fourney2024magenticone}, and \texttt{G-Designer}~\cite{zhang2025gdesigner}.
All methods are deployed on the same workers with 26 tasks across three departments.
We report API calls (API), sensitive span count (Spans), average span length (Len), prompt/completion tokens (P/C), and label diversity (Div., the number of span types with frequency at least three).

\noindent\textbf{Obs. \ding{175}: Structured organizational simulation improves the efficiency--richness trade-off of enterprise data generation.}
Table~\ref{tab:downstreamTask} shows that all methods generate internal-sensitive spans, but with different cost--quality profiles.
Magentic-One produces the largest number of spans and the highest label diversity, but at substantially higher API and token cost.
TaskWeave produces more spans than most baselines while using the fewest API calls and maintaining competitive span length and label diversity.
Compared with general-purpose frameworks such as AutoGen, MetaGPT, and CAMEL, TaskWeave yields richer enterprise-sensitive content under lower execution overhead.
On 300 short spans (length $\leq 10$), two advanced NER models~\cite{zhou2024uniner,ding2024rethinking} show F1 drops exceeding 40\%, indicating that OSSD differs substantially from standard public NER settings.

\subsection{Responsiveness and Transferability}
\label{sec:robust_transfer}

\textbf{Responsiveness.}
We evaluate external coupling in two directions: incident injection tests whether external events perturb organizational planning, while tool-enabled execution tests whether agents can write actions back to the environment.
We inject policy, economic, and technology incidents into Company A and equip selected agents with SQL, office, and communication tools.
Detailed event construction, injected examples, and tool assignments are provided in Appendix~\ref{app:environment}.

\begin{figure}[h]
\centering
\includegraphics[width=0.9\columnwidth]{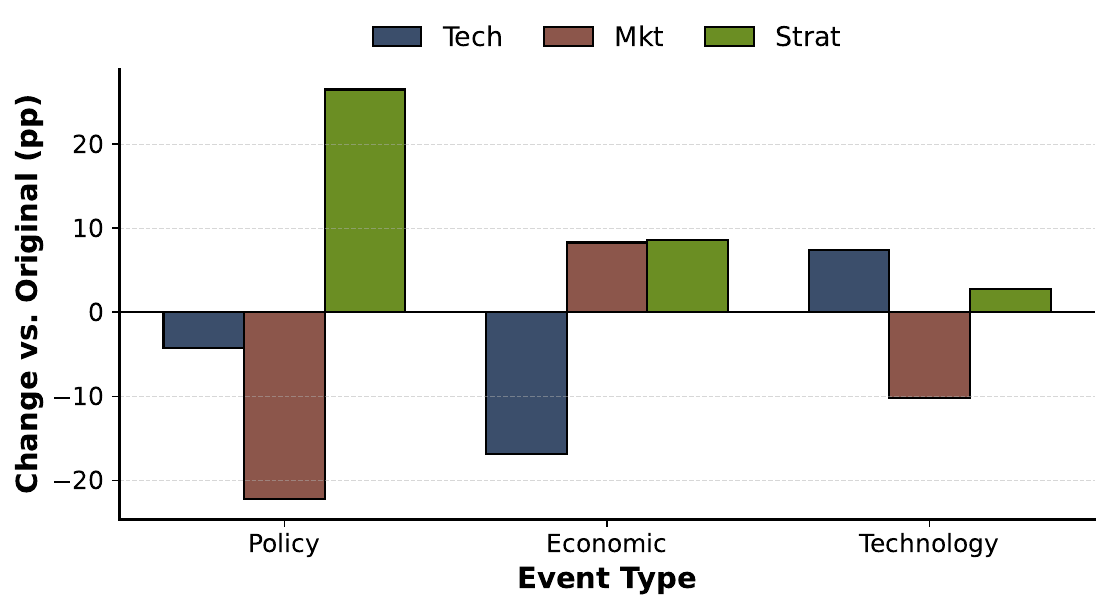}
\caption{Impact of external incidents on task allocation. Different events induce distinct shifts in strategic and operational workloads.}
\label{fig:external_incidents}
\vspace{-10pt}
\end{figure}

\noindent\textbf{Obs. \ding{176}: TaskWeave supports bidirectional environment interaction.}
Figure~\ref{fig:external_incidents} shows that injected incidents induce interpretable planning shifts: policy events increase compliance-oriented and strategic workloads, economic events strengthen cost-control and market-response activities, and technology events shift attention toward engineering robustness and product adjustment.
Tool-enabled agents further execute 534 tool calls, producing auditable documents, SQL queries, emails, and promotional posts.
These results show that TaskWeave can both adapt to external perturbations and write persistent actions back to the environment.

\textbf{Transferability.}
We instantiate TaskWeave in three additional organizations: a financial company (\textbf{Fin}), a manufacturing plant (\textbf{Manu}), and a government agency (\textbf{Gov}).
These settings vary in scale, hierarchy depth, and departmental structure, while using the same framework with re-instantiated organizational metadata.

\begin{table}[h]
\small
\caption{Cross-domain generalization. \textbf{Div.}: task diversity; \textbf{Role}: agent activations; \textbf{Com.}: completion rate.}
\setlength{\tabcolsep}{2pt}
\renewcommand{\arraystretch}{1.1}
\centering
\begin{tabular}{c|ccccccc}
\Xhline{1.2pt}
\rowcolor{CadetBlue!20}
\textbf{Scene} & \textbf{Size} & \textbf{Tiers} & \textbf{Dep.} & \textbf{Task} & \textbf{Div.} & \textbf{Role} & \textbf{Com.} \\
\Xhline{1.2pt}
Fin  & 15        & 3 & 3 & 20  & 45.05 & 119  & 81.6\% \\
\rowcolor{gray!20}
Manu & $\sim$50  & 4 & 4 & 129 & 54.36 & 655  & 88.0\% \\
Gov  & $\sim$100 & 5 & 5 & 290 & 58.55 & 1782 & 86.0\% \\
\Xhline{1.2pt}
\end{tabular}
\label{tab:generalizability}
\end{table}

\noindent\textbf{Obs. \ding{177}: Structural regularities transfer across domains.}
Table~\ref{tab:generalizability} shows that TaskWeave remains functional as organizational scale, hierarchy depth, and departmental complexity increase. Task diversity and agent activation grow with organizational complexity, while completion rates remain stable, suggesting that TaskWeave captures reusable organizational structure.

\section{Conclusion}
This paper presents \textit{TaskWeave}, an agentic framework for sustaining long-horizon organizational dynamics under explicit structural and contextual constraints.
At the core of TaskWeave is a multi-level simulation-memory mechanism that couples FPDA propagation with dependency-aware execution, allowing planning and execution states to remain coherent across temporal.
Experiments on year-long enterprise simulation, downstream artifact generation, environmental interaction, and cross-domain transfer show that TaskWeave supports structured, role-sensitive, and long-horizon organizational dynamics while producing grounded process outputs.
These results suggest that adaptive agentic simulation can serve as a controllable framework for studying long-horizon multi-agent dynamics, with enterprise modeling and data generation as important downstream benefits.

\section*{Limitation}
TaskWeave studies long-horizon organizational dynamics in controllable simulated enterprises. Although we evaluate year-long operation, external incidents, tool interaction, and cross-domain transfer, these settings remain simplified abstractions of real organizations. Future work can develop dedicated agent simulation environments with richer institutional rules, market signals, communication channels, and human-in-the-loop calibration. We also plan to extend current coherence-oriented evaluation with more standardized measures of behavioral and economic validity.

\bibliography{refrence}
\appendix
\clearpage

% Appendix-specific spacing. These do not change the ACL caption style.
\setlength{\textfloatsep}{8pt plus 1pt minus 2pt}
\setlength{\floatsep}{8pt plus 1pt minus 2pt}
\setlength{\intextsep}{8pt plus 1pt minus 2pt}

% Prompt box style.
% Requires: \usepackage[most]{tcolorbox} and \usepackage[table,dvipsnames]{xcolor}
\definecolor{mygrey}{RGB}{180,180,180}

\newtcblisting{promptbox}{
    listing options={
        basicstyle=\ttfamily\scriptsize,
        breaklines=true,
        breakatwhitespace=false,
        columns=fullflexible,
        keepspaces=true,
        showstringspaces=false
    },
    notitle,
    sharp corners,
    breakable,
    colframe=Periwinkle,
    colback=white,
    boxrule=1.2pt,
    boxsep=1pt,
    left=3pt,
    right=3pt,
    top=3pt,
    bottom=3pt,
    enhanced,
    shadow={2pt}{-2pt}{0pt}{opacity=0.6,mygrey},
    listing only
}

\section{Responsible Research and Open Resources}
\label{app:responsible}

\subsection{LLM Usage}
In preparing this work, we used large language models (LLMs) to assist with specific tasks, including linguistic refinement, grammar and style polishing, and related research exploration. All model outputs were carefully reviewed by the authors to ensure factual accuracy and faithfulness to the intended meaning of our work. We explicitly verified that no fabricated references, hallucinated claims, or misleading statements were introduced. The scientific content, analysis, and contributions remain solely the responsibility of the authors.

\subsection{Ethics Statement}
Our experiments involve the generation of synthetic enterprise data, which may contain elements resembling company backgrounds or personal information. We emphasize that all such data are entirely artificial and constructed for simulation purposes only. No real organizations, employees, or proprietary records are included, and the synthetic nature of the data ensures that it does not impact or compromise any real individuals or enterprises. We believe that this work poses no risk of privacy infringement or misuse of sensitive information.

\subsection{Open Resources}
Our code and generated data are available at:
\url{https://github.com/ZhuXuanCH/TaskWeave}.

\section{Additional Discussion}
\label{app:comparison}

\subsection{Comparison with BPM Methods}
\label{app:bpm-comparison}

Traditional enterprise modeling tools, such as \emph{Business Process Management Notation} (BPMN) simulators and process mining engines, have been widely adopted to model organizational workflows. However, these tools typically rely on static process templates and predefined control logic, making them less effective in environments that require continuous adaptation to shifting goals, unexpected events, or external shocks.

By contrast, TaskWeave complements such tools by enabling open-ended and hierarchical simulation through LLM-based agents. Compared with template- or log-driven approaches, TaskWeave provides: (i) higher fidelity by capturing both formal and informal coordination via dynamic intent propagation; (ii) greater flexibility by supporting adaptive task generation and multi-agent workflows without predefined process templates; and (iii) practical efficiency by reducing manual modeling effort while running on low-cost LLM backbones.

\begin{table}[t]
\centering
\small
\caption{Qualitative comparison between BPMN and TaskWeave.}
\label{tab:bpm-comparison}
\begin{tabular}{lcc}
\toprule
\textbf{Aspect} & \textbf{BPMN} & \textbf{TaskWeave} \\
\midrule
Process structure & Static & Dynamic \\
Adaptability & Low & High \\
Error handling & Predefined & Dynamic \\
Context integration & Limited & Supported \\
Domain portability & Limited & Supported \\
Simulation capability & Descriptive & Executable \\
\bottomrule
\end{tabular}
\end{table}

\subsection{Comparison with Existing MAS}
\label{app:mas-comparison}

Prior LLM-based MAS frameworks~\citep{hong2024metagpt,park2023stanfordtown} have demonstrated strong capabilities in collaborative task solving, sandboxed simulations, and meta-evaluation of generative agents. However, most of them emphasize local planning or isolated task execution, with limited centralized coordination or temporal hierarchy. As a result, they often fall short in modeling long-horizon enterprise workflows with continuity, dependencies, and adaptive feedback.

TaskWeave is not a generic MAS or dialogue-agent framework, but a structured system tailored for enterprise modeling and data synthesis. It uniquely integrates hierarchical planning, adaptive control loops, and traceable execution, enabling end-to-end simulation of organizational processes at scale. Table~\ref{tab:mas-comparison} presents a conceptual comparison.

\subsection{What Is Being Simulated?}
\label{sec:what_simulated}

Following recent agentic simulation work~\cite{park2023stanfordtown,gao2023s3}, our goal is not to claim that LLM agents literally reproduce human decision-making, but to study whether they can exhibit \emph{organizationally meaningful behavior} under structural constraints. TaskWeave is therefore best understood as a controllable simulator for studying long-horizon dynamics under structured constraints, rather than as a literal digital twin of a specific company. The empirical results suggest that explicit hierarchy, dependencies, and environmental conditions are sufficient for LLM agents to produce plausible coordination patterns, grounded process traces, and reusable simulation artifacts.

\subsection{Component Contribution Analysis}
\label{app:abalation}
A standard drop-one-component ablation is not well suited to TaskWeave because its core processes are tightly coupled and jointly form the simulation loop. We therefore analyze contribution through targeted evaluations of the framework’s core processes: planning-side state propagation (Sec.~\ref{sec:PropagationDynamics}) supports long-horizon coherence and plan continuity; structured coordination and role assignment (Sec.~\ref{sec:coherence}) improve delegation balance and workload coverage; and execution-side memory update (Secs.~\ref{sec:TaskExecution} and~\ref{sec:Downstream}) enables grounded execution with traceable outputs.

\subsection{Practical Cost and Scalability}
\label{app:cost}

In our setting, a fully fair end-to-end cost comparison against generic MAS frameworks is not directly comparable, since the available baselines do not support the same autonomous long-horizon planning-and-execution simulation.
We therefore report the measured cost of TaskWeave itself under the main setting. As shown in Table~\ref{tab:cost_discussion}, one complete year-long simulation of Company A costs \$17.84 under \texttt{GPT-4o-mini}, or \$1.49 when normalized monthly. The main expense does not come from isolated generations, but from maintaining \emph{organizational continuity} through long-horizon planning, repeated state updates, cross-role coordination, dependency resolution, and trace accumulation. Cost therefore scales with hierarchy depth, coordination density, replanning frequency, tool interaction, and organizational complexity.
\begin{table}[h]
\small
\caption{Practical cost of TaskWeave under the main Company A setting with \texttt{GPT-4o-mini}.}
\setlength{\tabcolsep}{3pt}
\renewcommand{\arraystretch}{1.1}
\centering
\begin{tabular}{c|ccc}
\Xhline{1.2pt}
\rowcolor{CadetBlue!20}
\textbf{Setting} & \textbf{Prompt} & \textbf{Completion} & \textbf{Cost} \\
\Xhline{1.2pt}
Full simulation & $\sim$77{,}300K & $\sim$10{,}300K & \$17.84 \\
\rowcolor{gray!20}
Monthly average (norm.) & $\sim$6{,}440K & $\sim$860K & \$1.49 \\
\Xhline{1.2pt}
\end{tabular}
\label{tab:cost_discussion}
\vspace{-10pt}
\end{table}

\subsection{On Realism and Evaluation Scope}
\label{app:grounding_discussion}

Our evaluation focuses on simulated organizations rather than the full historical record of a real enterprise, because real firms are difficult to study under controlled, repeatable conditions: internal decisions are often inaccessible, historical traces are incomplete, and counterfactual interventions are hard to evaluate systematically. We therefore position TaskWeave not as a literal digital twin, but as a controllable simulator for studying organizational process dynamics under structured settings. Under this framing, realism is judged less by one-to-one reconstruction than by whether the simulator exhibits meaningful organizational behavior. Our results support this view through linked execution traces (Sec.\ref{sec:TaskExecution}), process-grounded artifacts (Sec.\ref{sec:Downstream}), environmental responsiveness (Sec.\ref{sec:robust_transfer}), and transfer across organizational settings (Sec.\ref{sec:robust_transfer}). Existing multi-agent baselines remain only partially aligned with this goal, since they are typically designed for specific tasks or open-ended interaction rather than long-horizon simulation.

\begin{table*}[t]
\centering
\small
\setlength{\tabcolsep}{4pt}
\renewcommand{\arraystretch}{1.1}
\caption{Conceptual comparison of TaskWeave and prior LLM-based MAS frameworks. Symbols: \cmark\ supported, \xmark\ not supported, \pmark\ partially supported.}
\label{tab:mas-comparison}
\begin{tabular}{lcccccccc}
\toprule
\textbf{Framework} & \textbf{Sim} & \textbf{Hier} & \textbf{Goal} & \textbf{Long} & \textbf{Loop} & \textbf{Trace} & \textbf{Auto} & \textbf{Vers} \\
\midrule
\textbf{TaskWeave} & \cmark & \cmark & \cmark & \cmark & \cmark & \cmark & \cmark & \cmark \\
CAMEL              & \xmark & \xmark & \pmark & \xmark & \xmark & \xmark & \pmark & \cmark \\
AutoGen            & \xmark & \pmark & \cmark & \xmark & \xmark & \pmark & \cmark & \cmark \\
GPTSwarm           & \xmark & \pmark & \cmark & \xmark & \cmark & \pmark & \cmark & \cmark \\
MacNet             & \xmark & \pmark & \cmark & \xmark & \xmark & \pmark & \cmark & \cmark \\
GenAgents          & \pmark & \xmark & \xmark & \cmark & \xmark & \cmark & \cmark & \cmark \\
MetaGPT            & \pmark & \cmark & \cmark & \xmark & \pmark & \cmark & \cmark & \xmark \\
TwinMarket         & \cmark & \xmark & \xmark & \cmark & \xmark & \cmark & \cmark & \pmark \\
SOCIODOJO          & \cmark & \xmark & \pmark & \cmark & \pmark & \cmark & \cmark & \pmark \\
VIRSCI             & \pmark & \xmark & \pmark & \xmark & \xmark & \cmark & \cmark & \xmark \\
\bottomrule
\end{tabular}
\end{table*}

\section{Additional Evaluation Protocols}
\label{app:evaluation}

\subsection{Reliability of Evaluation Methods}
Given the novelty of our LLM-based MAS simulation setting, some open-ended evaluations necessarily involve human and LLM-based judgments. We followed best practices from prior MAS simulation works~\citep{park2023stanfordtown} and designed structured evaluation protocols to ensure rigor and transparency.

\paragraph{LLM-as-a-judge.}
As described in Section~\ref{sec:exp}, we designed prompts with three categories, nine criteria, and dual-descriptor definitions. We further validated reliability via Cohen's Kappa over 100 repeated samples, obtaining an agreement score of 0.88. In Section~\ref{sec:PropagationDynamics}, we implemented a two-stage pipeline consisting of generation and reasoning-based tracing, and verified consistency between two independent LLMs. For Section~\ref{sec:Downstream}, we applied a three-stage labeling pipeline.

\paragraph{Human expert evaluation.}
We involved four experts (A--D): Expert A has over 8 years of management experience in IT companies; Experts B--D have 1--2 years of experience in development or operations, all holding a master's degree or above. All experts underwent standardized training before conducting evaluations.

\begin{table*}[t]
\centering
\small
\renewcommand{\arraystretch}{1.15}
\caption{Evaluation methods and reliability measures.}
\label{tab:evaluation}
\begin{tabularx}{0.98\textwidth}{p{2.4cm} p{3.2cm} X}
\toprule
\textbf{Method} & \textbf{Location} & \textbf{Quality Assurance Strategy} \\
\midrule
LLM-as-a-judge & Section~\ref{sec:PropagationDynamics} & A two-stage pipeline is used for generation and reasoning-based tracing, with consistency checked across two independent LLMs. \\
LLM-as-a-judge & Section~\ref{sec:Downstream} & A three-stage labeling pipeline is applied for structured annotation and verification. \\
Human Expert & Sections~\ref{sec:simulation}/\ref{sec:coherence} & Expert A designed the role and task distributions as one realistic organizational configuration. \\
Human Expert & Section~\ref{sec:Downstream} & Experts B--D manually checked 500 samples; approximately 5--10\% label noise was retained to reflect LLM imperfection. \\
\bottomrule
\end{tabularx}
\end{table*}

\section{Agent Population and Organizational Settings}
\label{app:population}

\subsection{Role Construction}
To achieve a more realistic and credible simulation, we use customized prompts to define each agent by incorporating both personal identifiers, such as name, email, and education, and role-specific attributes, such as goals, responsibilities, and constraints. Representative role prompts are shown below.

\begin{promptbox}
MANAGERIAL ROLE PROMPT EXAMPLE

MANAGE_CPO = {
  "role": "Chief Product Officer",
  "organization": "PriGen",
  "responsibility": "Strategic decision-making, cross-department leadership, and alignment of product vision with business goals.",
  "biography": {
    "name": "Julia Townsend",
    "address": "352 Elm Acres, Markborough, FL 25220",
    "phone": "(582)-462-1236",
    "email": "karright@prigen.com",
    "education": "Stanford University, Master of Strategic Leadership"
  },
  "skills": [
    "Strategic Planning",
    "Cross-functional Leadership",
    "Communication",
    "Change Management",
    "Stakeholder Management"
  ],
  "goals": [
    "Align business goals with product strategy",
    "Improve team activation, tools, and autonomy",
    "Optimize cross-department collaboration and efficiency",
    "Lead PriGen through strategic growth phases"
  ],
  "constraints": [
    "Balance innovation with regulatory compliance",
    "Preserve operational continuity under uncertainty",
    "Operate under high visibility and performance expectations"
  ]
}

OPERATIONAL ROLE PROMPT EXAMPLE

DATA_ANALYST_CONFIG = {
  "role": "Data Analyst",
  "organization": "PriGen",
  "responsibility": "Analyze operational and strategic datasets, extract insights, and support data-driven decisions.",
  "biography": {
    "name": "Antonio McDonald",
    "address": "4522 Marcy Center Apt. 258, Davisville, HI 02231",
    "phone": "868-356-4288",
    "email": "ajmc@prigen.com",
    "education": "Shanghai Jiao Tong University, Master of Data Analytics"
  },
  "skills": [
    "Analytical thinking",
    "Problem solving",
    "Communication",
    "Impact management",
    "Adaptability"
  ],
  "goals": [
    "Deliver high-quality analytic outputs",
    "Contribute to strategic decisions",
    "Ensure timely execution of analytic tasks"
  ],
  "constraints": [
    "Comply with data security and organizational policies",
    "Coordinate with multiple stakeholders",
    "Balance technical depth with timely delivery"
  ]
}
\end{promptbox}

\subsection{Ideal Role Distribution}
\label{app:role-distribution}

We invite experts to design roles according to the simulated company background. The resulting ideal role distribution is shown in Table~\ref{tab:ideal-role-distribution}.

\begin{table}[t]
\centering
\small
\renewcommand{\arraystretch}{1.1}
\caption{Expert-defined ideal role distribution for the simulated SaaS enterprise.}
\label{tab:ideal-role-distribution}
\begin{tabular}{lc}
\toprule
\textbf{Role} & \textbf{Proportion} \\
\midrule
Product Manager & 0.15 \\
Marketing Specialist & 0.15 \\
Data Analyst & 0.15 \\
HR & 0.10 \\
Customer Success Manager & 0.10 \\
Technical Support Engineer & 0.08 \\
Backend Engineer & 0.07 \\
QA Engineer & 0.05 \\
UI/UX Designer & 0.05 \\
Frontend Engineer & 0.05 \\
DevOps Engineer & 0.05 \\
\bottomrule
\end{tabular}
\end{table}

\section{Generated Task Examples}
\label{app:generated-task-examples}

\begin{figure*}[t]
\centering
\includegraphics[width=0.82\textwidth]{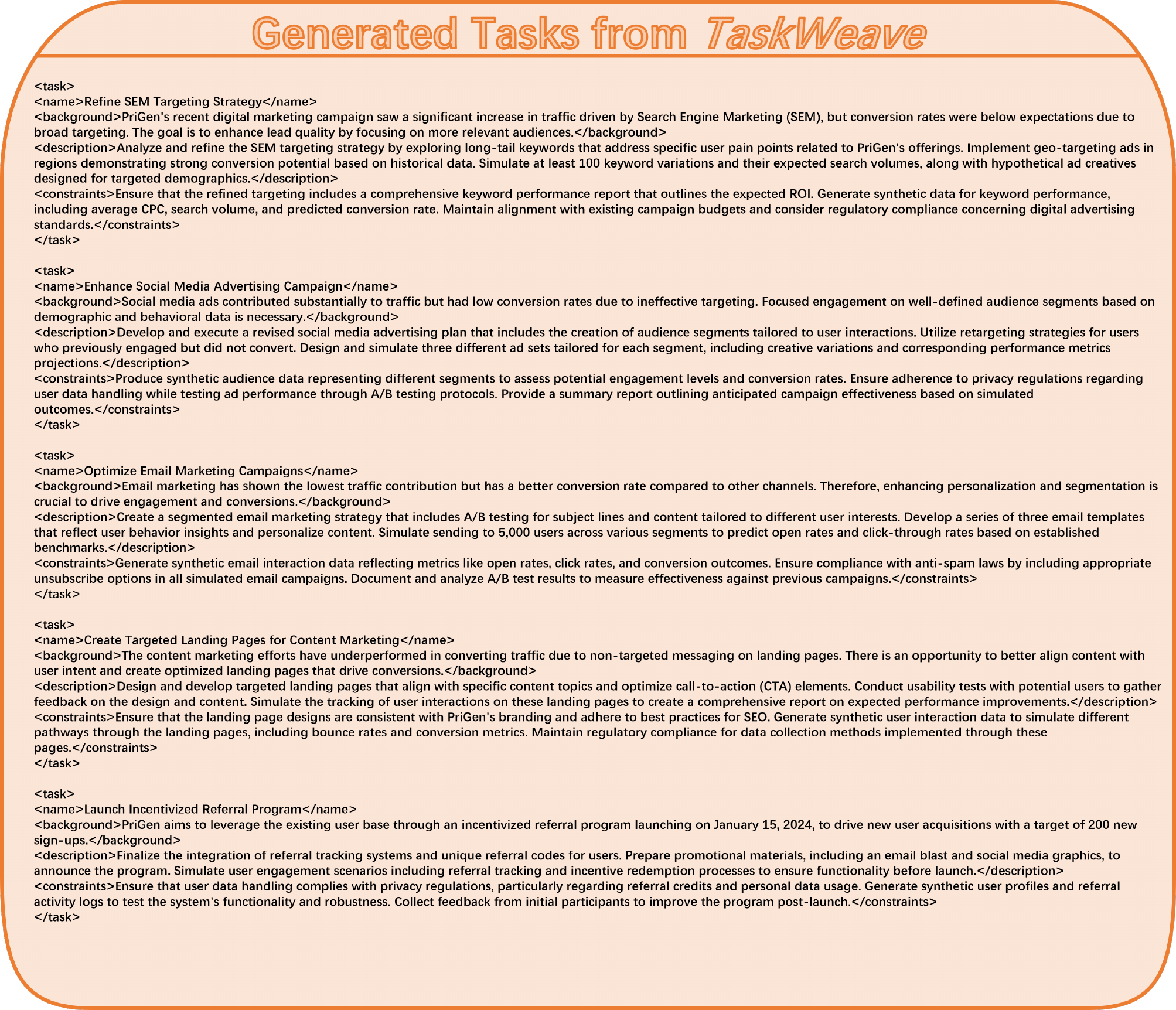}
\caption{Generated tasks from TaskWeave using ChatGPT-4o-mini.}
\label{fig:our-tasks}
\end{figure*}

Figure~\ref{fig:our-tasks} shows representative generated tasks from TaskWeave. TaskWeave organizes tasks in a structured and coherent manner, reducing duplication and better reflecting real-world workflow dependencies.

\section{Internal Text Hierarchical Classification}
\label{app:ithc}

Given a natural language input, the goal of Internal Text Hierarchical Classification (ITHC) is to assign it a fine-grained internal category from a predefined three-level taxonomy and to generate a textual rationale for the prediction. Let $\mathcal{T}$ denote the input-text space, $\mathcal{Y}$ the hierarchical label space, and $\mathcal{R}$ the rationale space. The task is to learn:
\[
f_{\mathrm{ITHC}}(T \mid \mathcal{D}_{\mathrm{gen}}) = (y,r),
T \in \mathcal{T},\ y \in \mathcal{Y},\ r \in \mathcal{R}.
\]
Each label $y=(y^{(1)},y^{(2)},y^{(3)})$ corresponds to a structured taxonomy level: category, subcategory, and fine-grained label. The explanation $r$ describes why the label applies to the input.

\subsection{Classification Label System}

To simulate enterprise dynamics in a realistic and controllable manner, it is crucial to ground agent behaviors in structured organizational knowledge. We propose a three-level hierarchical taxonomy tailored for internal enterprise documentation. The taxonomy supports task decomposition, trace explainability, cross-role collaboration, and fine-grained data generation and evaluation.

\begin{table*}[t]
\centering
\scriptsize
\setlength{\tabcolsep}{4pt}
\renewcommand{\arraystretch}{1.1}
\caption{Internal Text Hierarchical Classification label system.}
\label{tab:ithc}
\begin{tabularx}{0.98\textwidth}{p{2.7cm} p{3.2cm} X}
\toprule
\textbf{Primary Category} & \textbf{Secondary Category} & \textbf{Tags} \\
\midrule
Customer \& Marketing & Campaigns \& Promotions & MKT\_CAMPAIGN\_ANALYSIS; MKT\_CAMPAIGN\_PERFORMANCE; MKT\_EVENT\_OPERATION; MKT\_PROMOTION\_ANALYSIS \\
Customer \& Marketing & Customer Insights \& Analytics & CUST\_ENGAGEMENT\_REPORT; CUST\_FEEDBACK\_ANALYSIS; CUST\_MARKET\_ANALYSIS; MKT\_DATA\_REPORT \\
Customer \& Marketing & User Growth \& Profiling & MKT\_USER\_ACQUISITION; MKT\_USER\_CONVERSION; MKT\_USER\_PROFILE \\
Content \& Media & Content Creation \& Publications & CONTENT\_BLOG\_POST \\
Data \& Technology Management & Data Quality \& Infrastructure & DATA\_COMPLETENESS\_REPORT; DATA\_QUALITY\_ISSUE; OPS\_SYSTEM\_MONITORING \\
Data \& Technology Management & Technology Research \& Planning & OPS\_NEW\_TECH\_RESEARCH; STRAT\_TECH\_INNOVATION\_PLAN; STRAT\_TECH\_PARTNERSHIP\_REPORT \\
Human Resources & Talent Acquisition \& Onboarding & HR\_NEW\_EMPLOYEE\_REPORT; HR\_RECRUITMENT\_PLAN; HR\_RECRUITMENT\_RECORDS \\
Human Resources & Employee Development \& Training & HR\_TRAINING\_FEEDBACK; HR\_TRAINING\_PROGRAM; HR\_TRAINING\_RECORDS \\
Human Resources & Engagement \& Compliance & HR\_EMPLOYEE\_ENGAGEMENT\_REPORT; HR\_EMPLOYEE\_FEEDBACK; HR\_COMPLIANCE\_REPORT; HR\_POLICY\_DOCUMENT \\
Operations & Task Execution \& Management & OPS\_TASK\_EXECUTION; TASK\_EXECUTION\_STATUS; TASK\_EXECUTION\_SUMMARY \\
Security \& Compliance & Risk \& Policy Management & SEC\_COMPLIANCE\_AUDIT; SEC\_DATA\_PROTECTION\_GUIDELINE; SEC\_POLICY\_DOCUMENT; SEC\_INCIDENT\_RESPONSE\_PLAN \\
Strategy \& Innovation & Strategic Execution \& Growth & STRAT\_CROSS\_DEPARTMENT\_COLLAB; STRAT\_IMPLEMENTATION\_PLAN; STRAT\_USER\_EXPERIENCE\_IMPROVEMENT; STRAT\_USER\_GROWTH\_PLAN; STRAT\_WEBINAR\_IMPROVEMENT\_STRATEGY \\
Strategy \& Innovation & Market \& User Strategy & STRAT\_MARKETING\_STRATEGY; STRAT\_USER\_ENGAGEMENT\_STRATEGY \\
User Experience \& Research & Behavior \& Feedback Analysis & UX\_INTERACTION\_ANALYSIS; UX\_USER\_BEHAVIOR; UX\_USER\_ENGAGEMENT; UX\_USER\_FEEDBACK\_SUMMARY \\
User Experience \& Research & User Testing \& Research & UX\_USER\_RESEARCH\_REPORT; UX\_USER\_TESTING\_REPORT \\
\bottomrule
\end{tabularx}
\end{table*}

\subsection{Evaluating Model Outputs with ITHC}

To evaluate the consistency and semantic richness of document labels generated by different models, we propose a unified re-classification framework based on a three-stage self-consistency filtering pipeline using \texttt{GPT-4o-mini}. The prompt used for ITHC is shown below.

\begin{promptbox}
ITHC CLASSIFICATION PROMPT

AGENT ROLE:
You are the Data Protection Officer (DPO) at PriGen, a major telecom enterprise. You are responsible for classifying operational documents based on their content sensitivity, ensuring compliance with privacy, security, and regulatory standards.

OBJECTIVE:
Analyze and categorize documents into a hierarchical classification system, marking those that contain privacy-sensitive, legally protected, or confidential business data.

CLASSIFICATION SYSTEM:
The classification follows a multi-level structure, starting with broad categories and refining into more specific subcategories. Categories focus on customer data, employee records, financial transactions, telecom network information, security reports, and legal compliance.
{classification_system}

TASK INSTRUCTIONS:
1. You will receive a document from enterprise operations.
2. Analyze the document and classify it using the hierarchical classification system.
3. Assign the most specific applicable label(s) to the document.
4. If a document does not fit an existing label, propose a new label under the appropriate category.
5. Output the classification result in LIST format.
6. Avoid adding extra commentary or explanation outside the final classification output.
7. Strictly output in the expected format.
8. Multiple labels are allowed when necessary.

EXPECTED OUTPUT:
("Category > Subcategory > Label", "reason")

Where "reason" is a short explanation of why the document fits this label.

DOCUMENT TO BE CLASSIFIED:
{file_content}
\end{promptbox}

\begin{promptbox}
ITHC REVIEW PROMPT

AGENT ROLE:
You are a Classification Review Agent at PriGen. Your responsibility is to review and validate existing classification labels assigned to enterprise operational documents. Your goal is to ensure that each label is accurate, precise, and contextually appropriate.

TASK INSTRUCTIONS:
1. You will receive a document and its initial classification result.
2. Review whether each label is broadly reflective of the document content.
3. If a label is clearly wrong, misleading, or unrelated, replace it with a more suitable one.
4. If a label is generally acceptable, even if not perfect, retain it.
5. Be conservative in making changes and minimize revisions unless strongly justified.
6. Maintain the existing label hierarchy and structure.
7. Output the revised classification result in LIST format.
8. Do not include extra comments outside the required format.

EXPECTED OUTPUT:
("Category > Subcategory > Label", "reason")

DOCUMENT TO BE REVIEWED:
{file_content}

ORIGINAL LABELS:
{original_labels}
\end{promptbox}

In the first stage, \texttt{GPT-4o-mini} assigns a top-level category based solely on the original model output. In the second stage, it self-reviews the same input with access to its first-stage prediction and reasoning. In the third stage, it conducts another self-assessment based on the second-stage output. A label is accepted only if it remains unchanged across all three stages.

\begin{figure*}[t]
\centering
\includegraphics[width=0.95\textwidth]{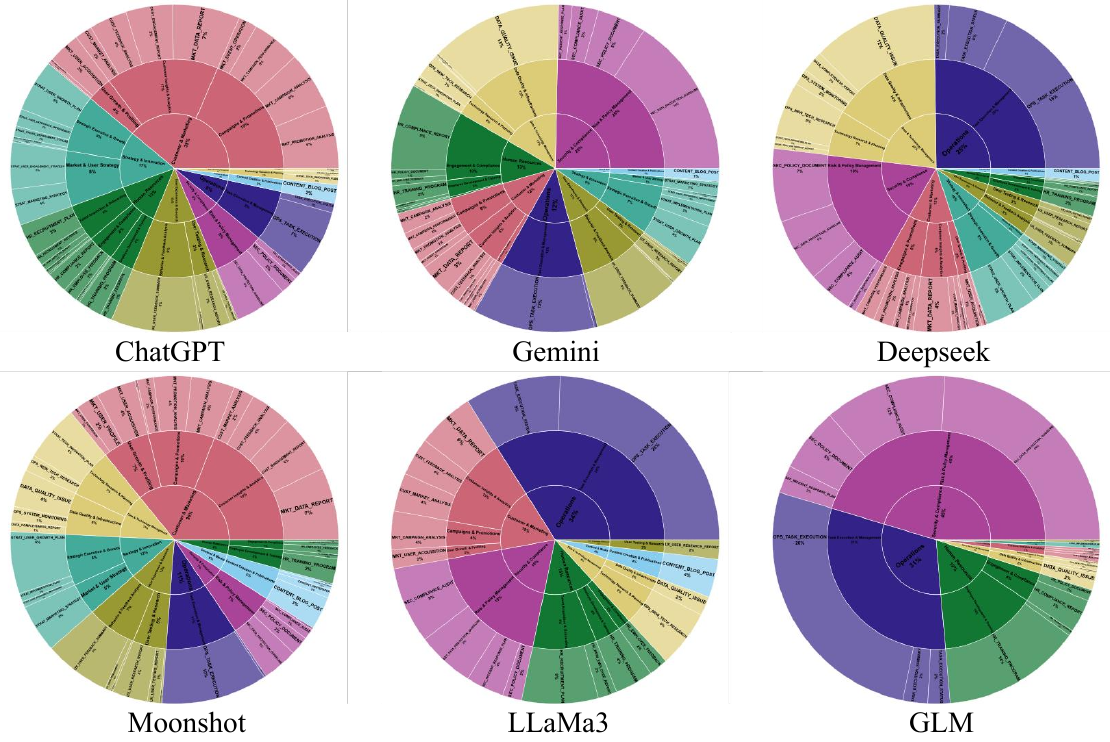}
\caption{ITHC outputs of six models.}
\label{fig:ithc}
\end{figure*}

This approach enables consistent comparison across six models: \texttt{GPT-4o-mini}, \texttt{Gemini-2.0-Flash}, \texttt{Deepseek-v3}, \texttt{Moonshot-v1-8K}, \texttt{LLaMA3.1-70B}, and \texttt{GLM-4-Flash}. As shown in Figure~\ref{fig:ithc}, the most frequent label across all models is \texttt{Customer \& Marketing}. However, semantic coverage varies considerably across models.

\section{MAS Checkpoint}
\label{app:MASCheckpoint}

This section describes the MAS Checkpoint mechanism, which serves as the backbone for task tracking and completion verification in TaskWeave. The mechanism simulates enterprise-grade planning and execution cycles by combining language-model-based judgment with structured sliding-window tracking.

\subsection{Task Pool Construction}

For each evaluation week $t$, we define a unified task pool $\mathcal{T}_t$ composed of newly planned tasks and previously uncompleted tasks from a retrospective window of $w$ weeks:
\[
\mathcal{T}_t = T_t \cup \bigcup_{i=1}^{w} T_{t-i}^{(u)} .
\]
Here, $T_t$ is the set of newly generated tasks for week $t$, $T_{t-i}^{(u)}$ denotes uncompleted tasks from week $t-i$, and $w$ is empirically set to 4 to match common enterprise monthly cycles.

\subsection{LLM-Based Task Decomposition}

To convert free-form planning text into structured task lists, we prompt a language model to act as a task-planning expert. The output is constrained to a Python list of clear and actionable items. We avoid requesting a fixed number of tasks; instead, we guide the model through prompt structure, examples, and abstraction control.

\begin{promptbox}
TASK DECOMPOSITION PROMPT

You are a precise task planning expert. Your responsibility is to transform a given project plan into a clear and actionable to-do list.

GUIDELINES:
- Read the entire plan carefully.
- Choose an appropriate level of abstraction.
- Tasks must originate from the plan content.
- Begin each task with an action verb.
- Output only a valid Python list.
- Do not include explanations, numbering, or extra commentary.

OUTPUT EXAMPLE:
["Design backend service", "Implement authentication", "Analyze user feedback", "Prepare deployment pipeline"]

PROJECT PLAN:
{plan_content}
\end{promptbox}

\subsection{Dual-Model Task Completion Evaluation}

Each task $\tau \in \mathcal{T}_t$ is evaluated independently by two LLMs: Gemini 2.5 Flash and GPT-4o-mini. These models determine whether the task is \texttt{Completed} or \texttt{Uncompleted} based on current documentation $C_t$:
\[
f_{\mathrm{LLM}}(\tau,C_t) \rightarrow
\{\texttt{Completed},\texttt{Uncompleted}\}.
\]

\begin{promptbox}
TASK COMPLETION REASONING PROMPT

You are an expert task completion analyst. Determine whether the task has been reasonably completed based on the provided context.

EVALUATION FRAMEWORK:
1. Understand the task purpose.
2. Review the summaries and output files.
3. Consider a task completed if:
   a. The goal was initiated or achieved.
   b. Indirect outcomes fulfill the task's intent.
   c. Partial work clearly contributes to the task.
4. Mark the task uncompleted if there is no meaningful evidence of progress.

OUTPUT FORMAT:
Return exactly one of the following:
"Reasoning..., yes"
"Reasoning..., no"

TASK:
{task}

CONTEXT:
{context}
\end{promptbox}

\subsection{Sliding-Window Lifecycle Tracking}

If a task remains uncompleted, it is carried into the next week's evaluation pool, provided that it still falls within the window constraint:
$
\tau \in T_t^{(u)}
\Rightarrow
\tau \in \mathcal{T}_{t+1}
\quad
\text{if}
\quad
t-\texttt{created\_week}(\tau)<w .
$

\begin{figure}[t]
\centering
\includegraphics[width=\columnwidth]{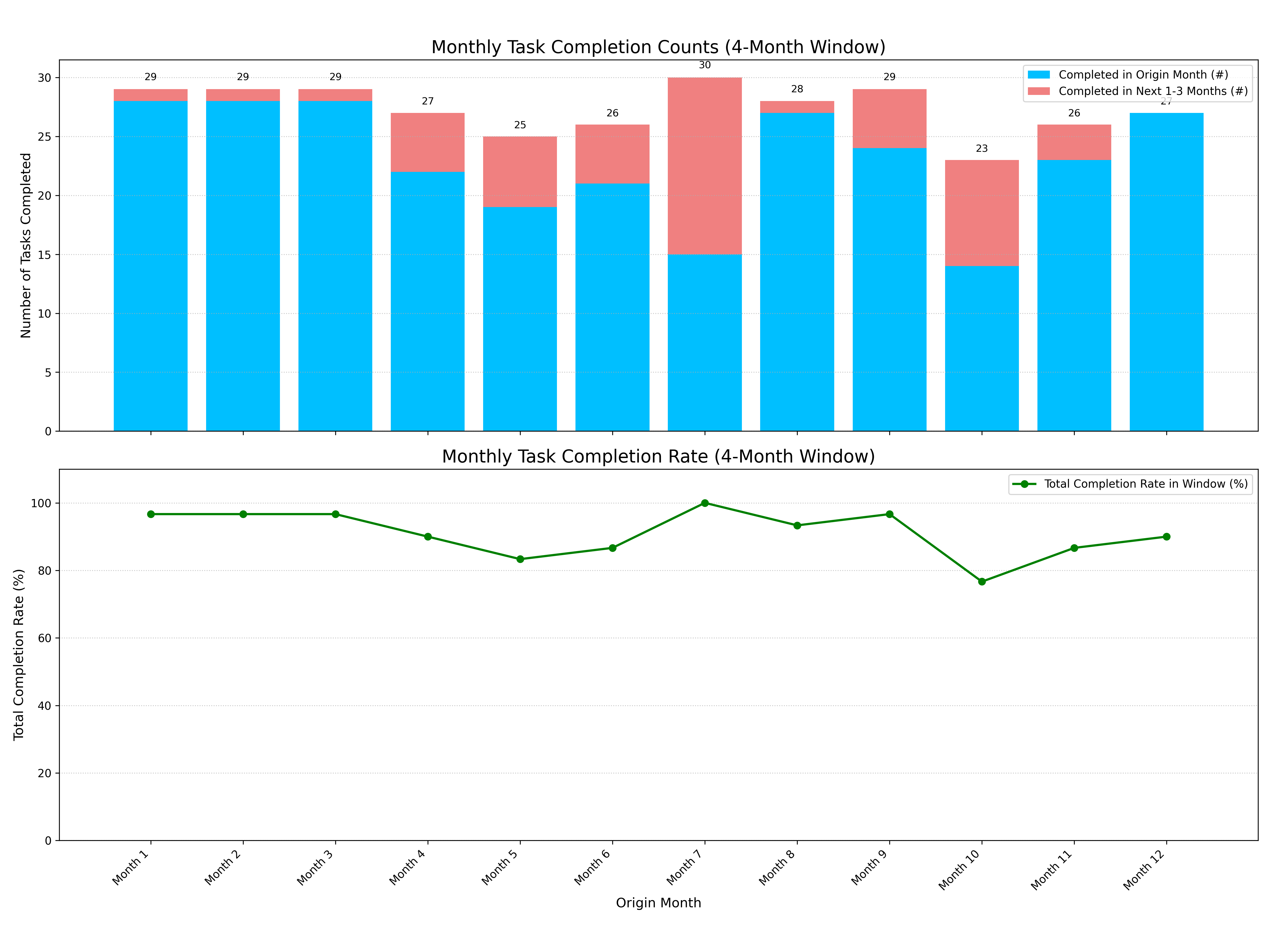}
\caption{Monthly completion counts and rates within a 4-month window.}
\label{fig:monthly-completion-subplot}
\end{figure}

\begin{figure}[t]
\centering
\includegraphics[width=\columnwidth]{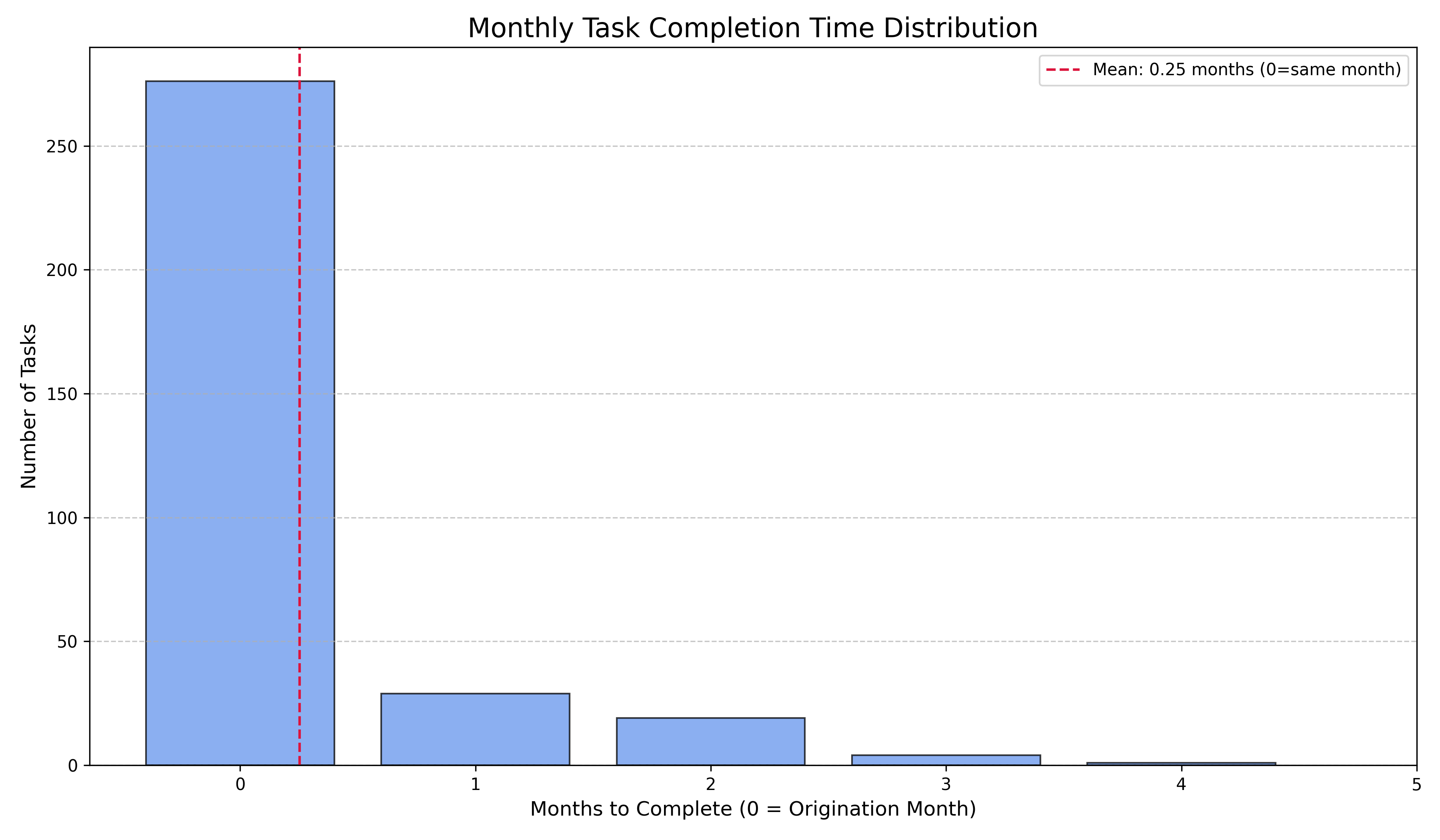}
\caption{Monthly completion time distribution.}
\label{fig:monthly-time-dist}
\end{figure}

As shown in Figure~\ref{fig:monthly-time-dist}, over 85\% of tasks are completed within the same month, indicating the effectiveness of LLM-generated decomposition in producing tractable workloads. Figure~\ref{fig:monthly-completion-subplot} further shows that tasks not completed in the origin month are typically recovered in subsequent months within the 4-month window.

To validate the reliability of LLM judgments, 10\% of evaluated tasks are randomly sampled and manually reviewed. If disagreement exceeds 10\%, full re-evaluation is conducted. This feedback loop also produces high-quality supervision data for potential model fine-tuning.

\subsection{Ablation Study and Completion Rate Analysis}

To evaluate the role of documentation $C_t$ in supporting model judgment, we conduct an ablation experiment. The original completion function is:
\[
\hat{y}_{\mathrm{full}}(\tau)=f_{\mathrm{LLM}}(\tau,C_t).
\]
The reduced variant is:
\[
\hat{y}_{\mathrm{ablated}}(\tau)=f_{\mathrm{LLM}}(\tau,C_t'),
\quad C_t' \subset C_t.
\]
The empirical completion-rate drop is:
\[
\Delta R = R_{\mathrm{full}} - R_{\mathrm{ablated}} \approx 50\%.
\]
Here, $C_t'$ excludes critical context such as status summaries or key execution lists. This substantial drop confirms the necessity of comprehensive context and validates the system's sensitivity to semantically grounded documentation.

\section{Organizational Sensitive Span Detection}
\label{app:OSSD}

Using LLMs to annotate or verify LLM outputs has proven effective in prior work on instruction generation~\citep{wang-etal-2023-self-instruct}, plan synthesis~\citep{liu2024apigen}, and trajectory validation~\citep{gao2025multimodal}. Building on these insights, we extend this paradigm to privacy annotation in simulated enterprise settings. We design a multi-stage structured prompting strategy for identifying and validating privacy-sensitive text spans, termed \textbf{Organizational Sensitive Span Detection} (OSSD).

Each annotation process contains two high-level stages. The \textit{label stage} identifies privacy-relevant content, extracts sensitive spans, assigns privacy category labels, and generates explanations. The \textit{check stage} re-evaluates the predicted labels and explanations for consistency, specificity, and semantic adequacy.

\subsection{Annotation Pipeline Overview}

The OSSD procedure consists of three stages: broad discovery, contextual refinement, and reasoning-based validation. The prompts used in these stages are shown below.

\paragraph{Stage 1: Broad Discovery.}
The model extracts a wide range of potentially sensitive spans. The prompt prioritizes recall and avoids committing to specific type assignments. The model returns:
\[
\mathcal{B}_x=\{(e_i,\texttt{``UNCERTAIN''})\}_{i=1}^{k},
\]
where $e_i$ is a candidate privacy-relevant span.

\begin{promptbox}
OSSD STAGE 1: BROAD DISCOVERY

Extract a list of potentially privacy-sensitive spans from the following document.

REQUIREMENTS:
- Each span should be returned as a tuple: (entity, "UNCERTAIN").
- Prioritize recall; include potentially sensitive business, employee, customer, operational, financial, or compliance-related spans.
- Do not assign specific privacy labels in this stage.
- Do not include explanations or label types beyond "UNCERTAIN".
- Output only a Python-style list of tuples.

DOCUMENT:
{file_content}
\end{promptbox}

\paragraph{Stage 2: Contextual Refinement.}
The system re-evaluates each span from Stage 1 and assigns a type $t_i \in \mathcal{T}$ from a predefined taxonomy, with a natural language explanation $r_i$:
\[
\mathcal{S}_x=\{(r_i,e_i,t_i)\}_{i=1}^{n}.
\]

\begin{promptbox}
OSSD STAGE 2: CONTEXTUAL REFINEMENT

You are a privacy analyst. Based on the initial candidate spans and the document context, refine the spans and assign each a specific privacy type from the taxonomy.

TASK:
- Review the document and the initial candidate spans.
- Remove irrelevant or overly generic spans.
- Refine span boundaries when necessary.
- Assign each retained span a specific privacy type.
- Provide a concise reason for each annotation.
- Return your answer as a list of (reason, entity, type) triples.

DOCUMENT:
{original_text}

INITIAL CANDIDATES:
{initial_labels_str}

PRIVACY TAXONOMY:
{privacy_taxonomy}
\end{promptbox}

\paragraph{Stage 3: Reasoning-Based Validation.}
The validation module audits Stage 2 outputs and revises or discards weak, vague, or inconsistent annotations:
\[
\mathcal{S}'_x=\texttt{Validate}(\mathcal{S}_x).
\]

\begin{promptbox}
OSSD STAGE 3: REASONING-BASED VALIDATION

You are performing a final consistency audit for privacy span annotations.

TASK:
Given the document and a list of (reason, entity, type) annotations, check whether each annotation is sufficiently precise and consistent with the assigned type.

VALIDATION RULES:
- Keep only valid (reason, entity, type) triples.
- Remove annotations whose entity is not actually sensitive in context.
- Remove annotations with vague, unsupported, or inconsistent reasoning.
- Correct obvious type mismatches when the correct type is clear.
- Do not invent spans that are not supported by the document.
- Output only the validated list.

DOCUMENT:
{original_text}

ANNOTATIONS TO VERIFY:
{phase2_labels_str}
\end{promptbox}

\subsection{Prompt Engineering and Stability}

To improve robustness, all stages use the same model, GPT-4o-mini, but with different temperatures: $T=0.8$ for Stage 1, $T=0.5$ for Stage 2, and $T=0.3$ for Stage 3. This setup introduces soft redundancy while stabilizing the final validation. We further evaluate consistency using Jaccard similarity:
\[
\mathrm{Jaccard}(\mathcal{S}_x,\mathcal{S}'_x)
=
\frac{|\mathcal{S}_x \cap \mathcal{S}'_x|}
{|\mathcal{S}_x \cup \mathcal{S}'_x|}.
\]
A match is defined on both span and type pairs $(e_i,t_i)$.

\subsection{Role Simulation as DPO}

This three-stage structure mirrors real-world workflows of a Data Protection Officer (DPO): the DPO first flags potentially sensitive content, then audits policy compliance and annotation quality, and finally checks logical consistency. By embedding explanatory rationales and enabling correction, OSSD supports transparency and interpretability in enterprise-level privacy annotation.

\begin{figure}[t]
\centering
\includegraphics[width=\columnwidth]{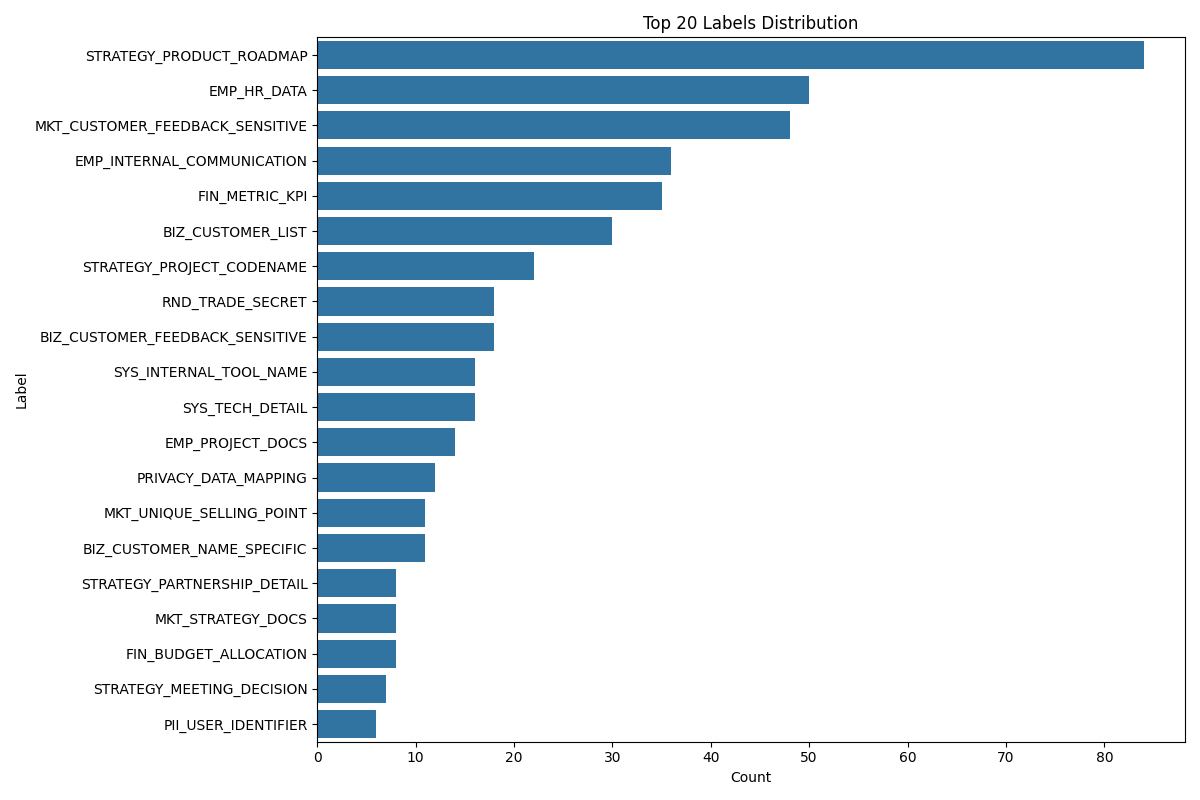}
\caption{Label distribution in OSSD.}
\label{fig:labeldist}
\end{figure}

\section{Generalizability}
\label{app:generalizability}

\subsection{Financial Company}
We evaluate TaskWeave by simulating the month-long operation of a representative financial company, denoted as Fin. The organization is composed of 15 role-specialized agents distributed across three organizational tiers.

\paragraph{Tier 1.}
Chief Executive Officer.

\paragraph{Tier 2.}
Chief Investment Officer, Chief Risk Officer, Chief Operations Officer.

\paragraph{Tier 3.}
Equity Trader, Fixed Income Analyst, Portfolio Assistant, Credit Risk Analyst, Market Risk Specialist, Internal Auditor, Settlement Officer, Compliance Associate, Fund Accountant, Relationship Manager, Client Onboarding Specialist.

\subsection{Manufacturing Company}
We evaluate TaskWeave by simulating the month-long operation of a representative automotive manufacturing company, denoted as Manu. The organization is composed of more than 30 role-specialized agents distributed across four organizational tiers.

\paragraph{Tier 1.}
Chief Executive Officer.

\paragraph{Tier 2.}
Chief Product Officer, Chief Marketing Officer, Chief Operations Officer, Chief Technology Officer.

\paragraph{Tier 3.}
Production Manager, Quality Manager, Maintenance Manager, Logistics Manager, Digital Systems Manager.

\paragraph{Tier 4.}
Under the Production Manager: CNC Operator, Tool Setter, Assembly Worker, Soldering Technician, Robotics Technician. Under the Quality Manager: Final Inspector, Visual Inspector, Compliance Auditor. Under the Maintenance Manager: PLC Technician, Wiring Technician, Machine Repairman, Hydraulics Technician, Predictive Maintenance Engineer. Under the Logistics Manager: Receiving Clerk, Forklift Operator, Shipping Coordinator, Boxing Technician, Packaging Design Engineer. Under the Digital Systems Manager: MES Analyst, Data Security Officer, AI Vision Specialist, Traceability Systems Specialist.

\subsection{Government Agency}
We evaluate TaskWeave by simulating the month-long operation of a representative government agency, denoted as Gov. The organization is composed of around 100 role-specialized agents distributed across five organizational tiers.

\paragraph{Tier 1.}
Minister.

\paragraph{Tier 2.}
Director of Strategic Planning, Director of Public Infrastructure, Director of Social Programs, Director of Policy Analysis, Director of Inter-Agency Coordination.

\paragraph{Tier 3.}
Chief of Strategic Planning, Chief of Infrastructure Development, Chief of Social Programs, Chief Policy Analyst, Chief Administrative Officer.

\paragraph{Tier 4.}
Senior Planning Manager, Urban Strategy Manager, Infrastructure Project Manager, Public Health Program Manager, Social Policy Manager, Sustainability \& Environment Manager, Policy Research Supervisor, Regional Coordination Manager, HR \& Admin Manager, Budget \& Resource Allocation Manager.

\paragraph{Tier 5.}
Planning Officer, Municipal Data Analyst, Health Policy Assistant, Citizen Complaint Registrar, Administrative Clerk, Digital Records Clerk, Public Service Trainee, Construction Analyst, Infrastructure Surveyor, Strategic Affairs Associate, Stakeholder Liaison, Legal Compliance Officer, Public Opinion Analyst, Welfare Program Officer, Community Outreach Assistant, Procurement Assistant, Citizen Engagement Officer, Smart City Systems Technician, Site Inspector, Field Research Associate, Education Policy Aide, Audit Support Officer, Training Coordinator, Statistical Reporting Assistant, HR Specialist, Transportation Planner, Scheduling \& Logistics Clerk, Project Evaluation Assistant, E-Government Support Officer, Facility Oversight Technician.

\section{Key Tasks Generated by TaskWeave}
\label{app:KeyTasks}

To understand the structural importance of tasks within the company's yearly operations, we analyze the directed task dependency graph generated by the multi-agent simulation. Nodes represent atomic business tasks, and edges indicate task-level dependencies or references. In-degree measures how many other tasks rely on a given task, while out-degree reflects how many downstream tasks a task depends on or triggers.

For double-column readability, Table~\ref{tab:degree} reports the top 10 tasks by in-degree and out-degree. The complete task graph and full ranked list are included in the released resources.

\begin{table*}[t]
\centering
\scriptsize
\setlength{\tabcolsep}{4pt}
\renewcommand{\arraystretch}{1.08}
\caption{Top tasks by in-degree and out-degree in the task dependency graph.}
\label{tab:degree}
\begin{tabularx}{0.98\textwidth}{XrXr}
\toprule
\textbf{Top In-Degree Task} & \textbf{In} & \textbf{Top Out-Degree Task} & \textbf{Out} \\
\midrule
Conduct\_Keyword\_Research\_for\_SEM\_Campaign & 103 & Review\_and\_Optimize\_User\_Engagement\_Strategies\_based\_on\_Feedback & 59 \\
Monitor\_and\_Analyze\_Referral\_Program\_Performance & 98 & Monitor\_and\_Optimize\_Digital\_Marketing\_Campaigns & 24 \\
Analyze\_Performance\_of\_Referral\_Program\_After\_Launch & 88 & Engage\_in\_Social\_Media\_Campaign\_Analysis & 24 \\
Optimize\_Email\_Marketing\_Campaigns & 75 & Optimize\_Digital\_Marketing\_Campaigns\_Based\_on\_Performance\_Data & 24 \\
Launch\_PriGen\_Referral\_Rewards\_Program & 58 & Conduct\_Deep\_Dive\_Analysis\_of\_SEM\_Performance\_Metrics & 22 \\
Create\_Focus\_Groups\_for\_UI\_Improvement\_Feedback & 58 & Initiate\_Keyword\_Optimization\_for\_SEM\_Campaigns & 21 \\
Enhance\_Incentivized\_Referral\_Program & 53 & Revise\_and\_Enhance\_the\_Onboarding\_Feedback\_Mechanism & 21 \\
Launch\_Incentivized\_Referral\_Program & 52 & Enhance\_User\_Feedback\_Mechanism\_through\_Incentives & 20 \\
Enhance\_Social\_Media\_Advertising\_Campaign & 51 & Initiate\_Content\_Marketing\_Strategy & 19 \\
Refine\_SEM\_Targeting\_Strategy & 50 & Optimize\_User\_Feedback\_Collection\_Post-Onboarding & 19 \\
\bottomrule
\end{tabularx}
\end{table*}

\section{Interaction with the External Environment}
\label{app:environment}

Our work focuses on designing a realistic and generalizable multi-agent system that simulates enterprise operations. External-environment modeling is important but complex: different enterprises face different environments, the required granularity must be designed case by case, and different simulation goals impose different requirements. This paper therefore concentrates on high-quality and generalizable modeling of intra-enterprise dynamics, while supplementing experiments that verify bidirectional interaction with external environments.

Specifically, TaskWeave supports both \textbf{ENV$\rightarrow$MAS} interaction, where external events are injected into the organization, and \textbf{MAS$\rightarrow$ENV} interaction, where agents invoke tools to act upon the environment.

\subsection{External-Event Injection}
\label{app:environment-inject}

To verify that external events can influence TaskWeave, we inject realistic real-world-based incidents and assess agent awareness through structured interviews. We also extract keywords from newly generated planning documents to quantify the resulting strategic shift.

\subsubsection{Experimental Configuration}
The experiments are based on the CompanyA scenario and use 4o-mini as the backbone model. The simulation spans one quarter, with events injected during the quarterly planning stage. We collected 70 events across three domains: Policy, Economic, and Technology. Among them, 15 events were injected into the experiments. Representative events are shown in Table~\ref{tab:representative-events}.

\begin{table*}[t]
\centering
\small
\setlength{\tabcolsep}{5pt}
\renewcommand{\arraystretch}{1.1}
\caption{Representative injected external events.}
\label{tab:representative-events}
\begin{tabularx}{0.98\textwidth}{p{3.5cm} p{2.2cm} X}
\toprule
\textbf{Title} & \textbf{Category} & \textbf{Real-World Basis} \\
\midrule
Digital Resilience Certification Mandate & Policy & EU CRA political agreement reached on 30 November 2023; certification bodies began booking audits for 2025. \\
Regulatory Tightening & Policy & During the first week of PIPL enforcement in August 2021, many SaaS firms received urgent self-inspection notices. \\
Cross-Border Data Ban & Policy & China's Cybersecurity Law led to Apple iCloud China data migration in 2017. \\
Macro Downturn & Economic & China's Q2 2022 GDP growth reached 0.4\%, and multiple SaaS earnings calls cited budget freezes. \\
Interest-Rate Spike & Economic & The Fed's first 25 bp hike in March 2022 triggered a global SaaS de-rating. \\
CPI 8\% Inflation Shock & Economic & U.S. inflation reached high levels in June 2022, increasing cost pressure for enterprises. \\
WASM-First Web Frameworks & Technology & Based on trends in Qwik, WASM adoption in Figma-like editors, and Cloudflare Workers with UI-thread offloading. \\
Zero-Day Cascade & Technology & The Log4j crisis in December 2021 required urgent global patching. \\
LLM-Driven Collaboration Norm Shift & Technology & Based on integration trends around Mixtral, Claude 3, DBRX, and OpenDevin-style workflows. \\
\bottomrule
\end{tabularx}
\end{table*}

\subsubsection{Event Impact}
We extract keywords in the generated plans before and after event injection:
\begin{itemize}[leftmargin=1.4em]
    \item \textbf{Original:} search engine marketing, tiered support system, user acquisition strategies, predictive analytics, AI-driven features.
    \item \textbf{Policy:} lifecycle assessments, Corporate Sustainability Reporting Directive, Cross-Border Data Ban, stakeholder engagement, resource allocation challenges, compliance documentation.
    \item \textbf{Economic:} zero-based budgeting, operational cash flow management, performance metrics, user acquisition strategies, churn risk mitigation, consultative selling techniques.
    \item \textbf{Technology:} digital marketing initiatives, document loading optimization, AI feature development, operational robustness, zero-trust architecture.
\end{itemize}

Agents also exhibit situational awareness in structured interviews and generate corresponding strategic responses, indicating that TaskWeave supports adaptive and context-aware planning in response to external dynamics.

\subsection{Tool Invocation}
\label{app:environment-tool}

In TaskWeave, each agent can be initialized with a set of tools, allowing it to interact with real-world products and operate with the local operating system. These tools enhance the MAS's ability to interact with the external environment, enabling the system to exert real-world effects and produce more authentic task outcomes.

\subsubsection{Experimental Configuration}
The experiment adopts the CompanyA architecture. We equip selected agents with 63 tools spanning three categories: SQL, social media, and office tools. We then run 30 distinct tasks. The detailed tool assignment for each agent is listed in Table~\ref{tab:agent-tools}.

\begin{table}[t]
\centering
\small
\setlength{\tabcolsep}{3pt}
\renewcommand{\arraystretch}{1.05}
\caption{Agent--tool assignment in Company A.}
\label{tab:agent-tools}
\begin{tabular}{p{0.43\columnwidth}p{0.50\columnwidth}}
\toprule
\textbf{Agent} & \textbf{Equipped Tools} \\
\midrule
Backend Engineer & SQLite, MotherDuck \\
Data Analyst & MotherDuck, Excel \\
Product Manager & Word, Excel, Email \\
Marketing Specialist & Excel, Email, Twitter \\
Customer Manager & Word, Email, Twitter \\
DevOps Engineer & MotherDuck, Email \\
HR Manager & Word, Email \\
Technical Support & -- \\
UI/UX Designer & -- \\
Frontend Engineer & -- \\
QA Engineer & -- \\
\bottomrule
\end{tabular}
\end{table}

\subsubsection{Tool Impact}

\textbf{SQL.} When data-analysis agents invoked SQL tools, they retrieved critical information from the database. The database acted as a centralized shared memory for the MAS, enhancing collaborative decision-making. During the experiment, \textbf{33 SELECT queries} and 8 additional SQL operations were executed.
 
\textbf{Social Media.} Agents responsible for market promotion proactively posted tweets and sent e-mails to partners. Company-wide announcements were also distributed via e-mail, reproducing internal communication workflows. Throughout the experiment, the MAS published \textbf{8 promotional tweets} and sent \textbf{46 e-mails}.

These results show that agents in TaskWeave can influence the external environment through tool usage.
\end{document}